\documentclass{article}

\usepackage[preprint]{neurips_2022}
\usepackage{graphicx}
\usepackage{amsmath}
\usepackage{amsthm}
\setcitestyle{numbers,square,comma}
\usepackage{fancyhdr}
\usepackage[colorlinks,
            linkcolor=green,       
            citecolor=blue, 
]{hyperref}
\usepackage{algorithm}
\usepackage{algorithmicx}
\usepackage{algpseudocode}
\usepackage{tabu}
\usepackage{wrapfig,lipsum,booktabs}
\usepackage{caption}
\usepackage{subfigure}

\usepackage[utf8]{inputenc} 
\usepackage[T1]{fontenc}    
\usepackage{hyperref}       
\usepackage{url}            
\usepackage{booktabs}       
\usepackage{amsfonts}       
\usepackage{nicefrac}       
\usepackage{microtype}      
\usepackage{xcolor}         
\usepackage{array}

\newcommand{\thickhline}{%
    \noalign {\ifnum 0=`}\fi \hrule height 1pt
    \futurelet \reserved@a \@xhline
}
\newcolumntype{"}{@{\hskip\tabcolsep\vrule width 1pt\hskip\tabcolsep}}

\newcommand{\PreserveBackslash}[1]{\let\temp=\\#1\let\\=\temp}
\newcolumntype{C}[1]{>{\PreserveBackslash\centering}p{#1}}
\newcolumntype{R}[1]{>{\PreserveBackslash\raggedleft}p{#1}}
\newcolumntype{L}[1]{>{\PreserveBackslash\raggedright}p{#1}}

\newtheorem{assumption}{Assumption}[section]

\title{Quantification before Selection: Active Dynamics Preference for Robust Reinforcement Learning}

\author{
  Kang Xu\\
  Academy for Engineer \& Technology\\
  Fudan University, Shanghai, China\\
  \texttt{kangxu21@m.fudan.edu.cn} \\
  \And
  Yan Ma\\
  Academy for Engineer \& Technology\\
  Fudan University, Shanghai, China\\
  \texttt{20210860024@m.fudan.edu.cn} \\
  \And
  Wei Li\thanks{Corresponding author.}\\
  Academy for Engineer \& Technology\\
  Fudan University, Shanghai, China\\
  Ji Hua Laboratory, Foshan, China\\
  \texttt{fd$\_$liwei@fudan.edu.cn} \\
}

\begin{document}

\maketitle

\begin{abstract}
Training a robust policy is critical for policy deployment in real-world systems or dealing with unknown dynamics mismatch in different dynamic systems. Domain Randomization~(DR) is a simple and elegant approach that trains a conservative policy to counter different dynamic systems without expert knowledge about the target system parameters. However, existing works reveal that the policy trained through DR tends to be over-conservative and performs poorly in target domains. Taking inspiration from active learning, we wonder if active sampling of parameters from the predefined randomization range can enhance the generalization performance of agents. To operationalize this idea, we introduce Active Dynamics Preference~(ADP), which quantifies the informativeness and density of sampled system parameters. Instead of uniformly sampling parameters like DR, ADP actively selects system parameters with high informativeness and low density. We validate our approach in four robotic locomotion tasks with various discrepancies between the training and testing environments. Extensive results demonstrate that our approach has superior robustness for system inconsistency compared to several counterparts.
\end{abstract}

\section{Introduction}
\label{sec:intro}
Reinforcement Learning has achieved remarkable successes in video games \cite{ecoffet2021first, jaderberg2019human} and continuous robot control tasks\cite{levine2018learning, andrychowicz2020learning}. Recent advances have demonstrated real-world deployments of RL algorithms~\cite{peng2020learning, hwangbo2019learning}. However, the unknown disturbance in the practical systems might produce a model mismatch concerning the model utilized in simulation~\cite{yu2017preparing, peng2018sim}. The behavior of policies under the model inconsistency is unpredictable, which might result in catastrophic damage to the physical systems~\cite{ross2011reduction}. To this end, robust handling of the model uncertainties is a must before fully leveraging reinforcement learning algorithms in real-world applications.

One promising but simple approach to overcoming the unpredictable model discrepancies is the Domain Randomization~(DR)~\cite{peng2018sim, tobin2017domain}. With the parameterizable simulation environments, we define intervals for some physical parameters (torso mass, friction, etc...) in DR. At the beginning of each episode, we train the policy in the environment with the system parameters uniformly sampled from the predefined ranges, and we hope the final trained agent performs well in some target domains or behaves robustly when the system dynamics vary. However, recent work suggests that the policy trained via DR may appear over-conservative, thus making the agent performs poorly in target domains ~\cite{xie2021dynamics}. In addition, it is difficult to find proper parameter randomization ranges to train a robust policy when we have limited expert knowledge about the physical system.

In this work, we take a step toward resolving the robust policy training problem from the perspective of active task sampling~\cite{mehta2020active, toneva2018empirical, florensa2018automatic, zhang2020automatic, jiang2021prioritized}. The key insight of this paper is that the uniform sampling method adopted in the traditional DR approach might be ill-conditioned for the evolving policy in the training procedure. Systems with different parameters provide different levels of difficulty for the policy. For example, it is more difficult for a quadruped robot to walk when equipped with extra torso mass than in normal circumstances. At the early stage of policy training, challenging tasks will provide no effective learning signal for the unmatured agent. If we can actively sample systems with proper parameters during training, we might prevent the policy from over-conservative behavior and stabilize the training.

We introduce Active Dynamics Preference (ADP), a general and powerful method that tackles robust RL from the sampling perspective. To sample systems with appropriate parameters, we estimate the degree of informativeness via the temporal difference error of the trajectories sampled from the system. Besides, we record the visited counts of system parameters to prevent the policy from overfitting. However, the parameter ranges are continuous with uncountable implicit tasks. We propose a performance continuity hypothesis to discretize the predefined parameter ranges to several bins. Furthermore, we claim that novelty or informativeness should be adaptively combined during training. Thus we introduce a bandit to actively select the trade-off coefficient. ADP samples several system parameters from the predefined range and a small-capacity FIFO parameter replay buffer. Then ADP selects system parameters utilizing a preference for high informativeness and low density. After the policy update, ADP updates the corresponding table values, and inserts the selected parameters into the replay buffer. The buffer stores the most recent selected parameters to increase the visit frequency of the relatively more valuable parameters.

\begin{figure*}[t]
    \centering
    \includegraphics[width=0.95\textwidth]{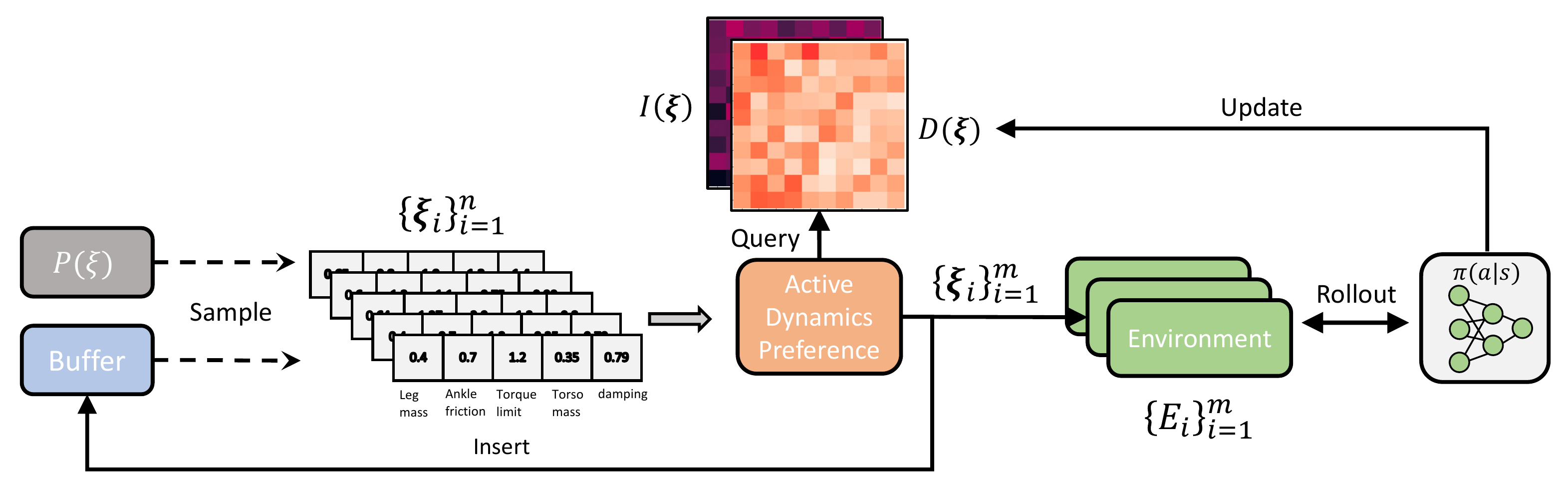}
    \vspace*{-1mm}
    \caption{Framework of Active Dynamics Preference~(ADP). We sample $n$ system parameters from the predefined randomization space $P(\xi)$ and a small-capacity FIFO parameter replay buffer at the beginning iteration. By querying the informativeness score function $I(\xi)$ and density function $D(\xi)$ which are both characterized as discrete tables in our work, we select $m~(m<n)$ parameters with the highest informativeness scores and the lowest density values. After the rollouts and policy training, we update the corresponding function values of the selected system parameters.}
    \label{fig:framework}
    \vspace*{-2mm}
\end{figure*}

In summary, our contributions are threefold: (1) we propose an efficient and general algorithm to enhance the generalizibility of agents trained via DR; (2) by testing algorithms in various system parameters, we reveal the existence of several system parameters which result in significant performance degradation, while these parameters are typically chosen for randomization in prior works; (3) we demonstrate that our approach improves the policy robustness in various model mismatch scenarios including modeled system parameter errors and unmodeled system discrepancies.

\section{Related Work}
\textbf{Active Sampling for Domain Randomization} 
Domain Randomization (DR) is a simple and effective method for training robust policies. However, the predefined training parameter ranges require iterative adjustment by trial and error. Several works have been proposed to improve DR by actively sampling informative tasks from the predefined task set~\cite{mehta2020active, mozian2020learning, rajeswaran2016epopt,ramos2019bayessim, muratore2021data}. Metha et al.~\cite{mehta2020active} proposed Active Domain Randomization (ADR) which actively samples tasks via particles trained by SVPG~\cite{liu2017stein}, and the particles are encouraged to sample tasks where the agent behaves differently compared to a reference environment. The behavior difference is quantified by a discriminator reward model similar to the prior work~\cite{eysenbach2018diversity}. ADR assumes there exist parameters that are more informative than others, and a reference environment is required for behavior comparison. The main drawback of ADR is the soundness of the assumption, which might not hold for different system parameters. Besides, the training for the discriminator model and particles requires extra computation redundancy. In contrast, we select system parameters and quantify the informativeness via TD-errors rather than extra modules. Rajeswaran et al.~\cite{rajeswaran2016epopt} proposed to optimize worst-case performance of the policy by utilizing conditional value at risk (CVaR)~\cite{tamar2015optimizing}. However, trajectories whose returns are lower than the risk threshold are discarded in practical, which leads to poor sample-efficiency. Bayesian optimization~\cite{snoek2012practical} has been introduced to adjust randomization space in DR to approach the parameter distribution in some specific target domain~\cite{ramos2019bayessim,muratore2021data}, which is different from our setting that we focus on unpredictable disturbance rather than a fixed target domain. 

\textbf{Robust Reinforcement Learning} 
To improve policy robustness regarding disturbances on the system parameters, several approaches have been proposed in recent years~\cite{pinto2017robust, tessler2019action, kamalaruban2020robust,vinitsky2020robust, jiang2021monotonic, wang2021online, eysenbach2021robust}. Pinto et al.~\cite{pinto2017robust} proposed Robust Adversarial Reinforcement Learning (RARL) to formulate the robust RL problem as a two-play zero-sum game. By modeling external forces as actions of an adversary, the robustness of the agent is improved by training both policies jointly. Tessler et al.~\cite{tessler2019action} proposed Probabilistic Action Robust MDP (PR-MDP) and Noisy Action Robust MDP (NR-MDP) to formulate robust RL problem. Eysenbach et al.~\cite{eysenbach2018diversity} demonstrated that information compression provides a straightforward mechanism for training robust policy. Our work is orthogonal to these approaches in that we address the policy robustness problem from the perspective of task sampling.

\textbf{Automatic Curriculum Learning for Deep RL}
Extending the curriculum learning to deep RL has demonstrated better sample-efficiency in hard tasks~\cite{baranes2009r, florensa2017reverse, ivanovic2019barc, matiisen2019teacher} or advanced resulting policy generality~\cite{portelas2020teacher, baker2019emergent, zhang2020automatic, jiang2021prioritized}. Especially in goal-conditioned RL, several works introduce curriculum to policy learning efficiency~\cite{andrychowicz2017hindsight, florensa2018automatic, racaniere2019automated, colas2019curious, fang2019curriculum, zhao2019curiosity}. Unlike existing works, we extend the curriculum learning to the problem of robust policy learning in the presence of perturbation over the system dynamics. While the robustness can be regarded as one category of the generality, different tasks in this setting lead to different transition functions in the MDPs, which is orthogonal to the goal-conditioned RL~\cite{andrychowicz2017hindsight} and unsupervised environment design~\cite{jiang2021replay}. In addition, the infinite tasks in our setting can be problematic for task sampling in practice. The most related work is PLR~\cite{jiang2021prioritized} that prioritizes the training levels from the procedural content generation games~\cite{cobbe2020leveraging} to improve the policy generalization, and we build our work on PLR. However, we focus on the generalization across various dynamics, which is critical to the Sim2Real deployment of RL algorithms in robotics research~\cite{peng2018sim,kumar2021rma,du2021auto}.

\section{Background}
\label{sec:bac}
\subsection{Notation}
We formulate our environment as a Markov Decision Process (MDP) \cite{puterman1990markov}, $\mathcal{M}=(\mathcal{S}, \mathcal{A}, \mathcal{P}_\xi, r, \gamma, \rho_0)$, where $\mathcal{S}$ is the state space, $\mathcal{A}$ is the action space, $r: \mathcal{S}\times\mathcal{A}\times\mathcal{S} \to \mathbb{R}$ is the reward function, $\rho_0$ is the initial state distribution, and $\gamma \in \left[0, 1\right]$ is the discount factor. In the context of systems with various physics parameters, we denote the physics parameters to $\xi \in \mathbb{R}^d$ where the $d$ is the numbers of parameters potentially get disturbed. We formulate the transition function as $\mathcal{P}_\xi : \mathcal{S}\times\mathcal{A}\to\mathcal{S}$ which is implicitly conditioned by the physics parameters $\xi$. We define the trajectory as $\tau = (s_0, a_0, s_1, a_1, \dots, s_T, a_T)$ where $T$ is the length of the trajectories and $a_i$ is sampled from the policy $\pi(a_t|s_t): \mathcal{S} \to \mathcal{A}$.

\subsection{Domain Randomization}
Domain Randomization (DR) aims to improve the average performance of the policy concerning various physics parameters. In order to overcome the potential disturbance of some system parameters or to approximate the system parameters in the target domain, a randomization space $\Xi = \{ (\xi^{(i)}_{low}, \xi^{(i)}_{high}) \}_{i=1}^d$ is defined. DR samples system parameters $\xi \in \Xi$ and generates the corresponding environment $E_\xi$. Then the policy $\pi_\theta(a_t|s_t)$ is optimized based on the trajectories sampled from the environment $E_\xi$. The objective of DR is to find a policy $\pi_\theta(a_t|s_t)$ that maximizes the policy performance on expectation over the randomization space $\Xi$:
\begin{equation}
    J(\pi_\theta) = \mathbb{E}_{\xi\sim \Xi}\left[
    \mathbb{E}_{
    \pi_\theta, \mathcal{P}_\xi, \rho_0
    } \left[
        \sum_{t=1}^T \gamma^t r_t
    \right]
    \right]
\end{equation}

\section{Active Dynamics Preference}
In this section, we propose ADP to obtain a robust policy from the perspective of active task sampling. The overview of ADP is described in Section~\ref{method:overview}. Other details of
our algorithm is described in Sections~\ref{method:quantify}-\ref{method:bandit}.

\subsection{Overview of ADP}
\label{method:overview}
At each iteration of ADP, we sample several system parameters $\{\xi_i\}_{i=1}^n$ from the predefined randomization space $\Xi$ and a parameter replay buffer $\mathcal{B}$. Instead of utilizing these parameters for rollouts straightly, we first measure the parameters from the perspective of informativeness and novelty. We utilize an informativeness score function $\mathcal{I}(\xi): \mathbb{R}^d \to \mathbb{R}$ and a density function $\mathcal{D}(\xi): \mathbb{R}^d \to \mathbb{R}$. After measurement of the sampled parameters, we select a subset of the sampled parameters $\{\xi_i\}_{i=1}^m$ via a preference for high informativeness scores and low-density values. The selected parameters then get passed to the simulator, and a set of environments $\{E_i\}_{i=1}^m$ are generated. Finally, the agent rollouts with the environments, and two functions get updated at the end of each iteration. Refer the pseudo code to Algorithm~\ref{alg:adp}.

\begin{algorithm}[b]
    \caption{Active Dynamics Preference}
    \label{alg:adp}
    
    {\bf Input:}
    Policy $\pi_\theta(a|s)$, informativeness score function $\mathcal{I}(\xi)$, density function $\mathcal{D}(\xi)$, predefined parameter randomization space $\Xi = \{ (\xi^{(i)}_{low}, \xi^{(i)}_{high}) \}_{i=1}^d$, parameter replay buffer $\mathcal{B}$, trade-off coefficient bandit, validation environment set $\{E\}^V$.
    
    {\bf Output:} Policy $\pi_\theta(a|s)$
    
    \begin{algorithmic}[1]
        \For{iteration = 1, 2, \dots}
            \State Sample $\{\xi_i\}_{i=1}^n$ jointly from $\Xi$ and $\mathcal{B}$.
        
            \State Sample trade-off coefficient $\omega$ from the bandit.
            
            \State Evaluate the parameters with $\mathcal{I}(\xi)$, $\mathcal{D}(\xi)$ and get two sets of scores $\{I_i\}_{i=1}^n$, $\{D_i\}_{i=1}^n$.
            
            \State Rank the scores: $I_i = rank(I_i), D_i = rank(D_i), i=1,\dots,n$.
            
            \State Select system parameters subset $\{\xi_i\}_{i=1}^m$ via Equation~\ref{eq:selection}. 
    
            \State Generate environments $\{E_i\}_{i=1}^m$.
            
            \State Generate trajectories $\{\tau_i\}_{i=1}^m $ from the environments.
            
            \State Update the corresponding values of $\mathcal{I}(\xi)$ and $\mathcal{D}(\xi)$ with the trajectories $\{\tau_i\}_{i=1}^m$.
            
            \State Training policy $\pi_\theta$ with the trajectories $\{\tau_i\}_{i=1}^m$ using PPO.
            
            \State Update the beta distribution of the bandit's arm with the performance of $\pi_\theta$ in $\{E\}^V$.
    
        \EndFor
    \end{algorithmic}
    
\end{algorithm}

\subsection{Dynamics Parameters Quantification}
\label{method:quantify}
We argue that uniformly sampled system parameters might provide inappropriate levels of difficulty for the policy. We aim to quantify the informativeness of different systems and actively sample the systems with proper parameters. One option is to use the cumulative reward sampled from the system. However, the policy gets updated based on the trajectories sampled from partial systems, and the performance in any system might change after single policy optimization. For example, the policy performance in one system with parameters $\xi_A$ might change after a single update with trajectories sampled from the dynamic system with parameters $\xi_B$. Taking inspiration from prior works which estimate the learning potential of different tasks~\cite{jiang2021prioritized,jiang2021replay}, we introduce Informativeness Score Function (ISF) based on the Generalized Advantage Estimation (GAE) \cite{schulman2015high} to measure the uncertainty of the system difficulty:
\begin{equation}
    \mathcal{I}(\xi) = \mathbb{E}_{\pi_\theta,P_\xi}\left[ \frac{1}{T}\sum_{t=1}^T 
    \left|
        \sum_{k=t}^T (\gamma\lambda)^{k-t}\delta_t 
    \right|
    \right]
\end{equation}

The GAE at timestep $t$ is the exponential discounted sum of all $T-t$ step TD-errors $\delta$ from t, which was proposed to control the trade-off between the bias and variance in the policy gradient methods~\cite{schulman2015trust, schulman2017proximal}. Here we utilize the absolute GAE value to approximate the value function prediction error of the samples. 

Utilizing the absolute GAE to evaluate the informativeness of tasks was first proposed in PLR~\cite{jiang2021prioritized}. PLR hypothesizes that the value targets of the tasks beyond the current ability of the agent tend to be stationary compared with those of tasks at the frontiers of the agent's ability. Thus, the corresponding GAE values are lower than those from tasks at the frontiers of the agent ability. We follow the hypothesis of PLR, while we apply the absolute GAE to evaluate the informativeness of tasks with various dynamics resulting from different physical parameters.


To prevent the policy from overfitting to a subset of all predefined systems, we introduce novelty measurement for the selected system parameters. Here we introduce the count-based method \cite{bellemare2016unifying} to record the visited counts of the parameters. We propose Density Function (DF) for the parameters:
\begin{equation}
    \mathcal{D}(\xi) = \frac{c_\xi}{\sum_{\hat{\xi}\in \Xi} c_{\hat{\xi}}}
\end{equation}
to evaluate the novelty of the parameters, where the $c_\xi$ is the number of times the parameter $\xi$ is selected.

To evaluate the systems concerning both informativeness and novelty, we need to handle the different scales of two types of scores. Here we propose rank transformation of the informativeness scores and novelty values: $I_i = rank(\mathcal{I}(\xi_i)), D_i = rank(\mathcal{D}(\xi_i)), i=1,\dots,n$.

\subsection{Adaptive Preference Correction}
\label{method:bandit}
Selecting the parameters requires a trade-off coefficient to control the preference of the two indicators above. However, the optimal trade-off coefficient may vary during the training process. For example, we might favor novel parameters in the early stage of the training to obtain essential evaluations across all systems. We formulate this problem through the lens of multi-armed bandits and adaptively select the trade-off coefficient from a predefined set. 

To favor informativeness or novelty at different stages of training by adaptively sampling the coefficient $\alpha$, we propose the trade-off coefficient selection as a multi-arm bandits and denote the candidates of trade-off coefficient $\Omega=\left\{\omega_1=0.2,~\omega_2=0.8\right\}$. Furthermore, we introduce Thompson Sampling~\cite{agrawal2012analysis,russo2016information} to select from the two arms. The reward model of each arm is initialized by a beta distribution with $\alpha_i^0 = \beta_i^0=1$:
\begin{equation}
    \mathrm{P}(r|\omega_i,\Theta^0) = \dfrac{\Gamma(\alpha_i^0+\beta_i^0)}{\Gamma(\alpha_i^0)\Gamma(\beta_i^0)} r^{\alpha_i^0-1}(1-r)^{\beta_i^0-1},
\end{equation}
where $i$ denotes the $i$th arm, the superscript denotes the iteration, $\Gamma(x)$ is the gamma function, and $\Theta^0$ denotes the prior parameters of the beta distributions.
At each iteration $t$, we sample candidates $\{r_i\}_{i=1}^2$ from the distributions, and pull the arm $i^t = argmax_{i\in\{1, 2\}}r_i$. Thus the corresponding coefficient $\omega_{i^t}$ is selected. The sampling process can be formulated as:
\begin{align}
    &{i^t} = \arg\max_{i\in\{1, 2\}}r_i
    \label{eq:prefrence_sampling}\\
    \text{where}~&~r_i \sim \mathrm{P}(r|\omega_i,\Theta^t),~i=1,2\nonumber
\end{align}

At the end of each iteration, we evaluate the policy in predefined validation environments (described in Section~\ref{exp:envset}) and obtain an evaluation return $R^t$. We update the parameters of the beta distribution of arm $i^t$ as follows:
\begin{equation}
    (\alpha_{i^t}^{t+1},\beta_{i^t}^{t+1}) = 
    \begin{cases}
      (\alpha_{i^t}^{t} + 1, \beta_{i^t}^{t}), & R^t > R^{t-1}\\
      (\alpha_{i^t}^{t}, \beta_{i^t}^{t} + 1), & R^t \leq R^{t-1}
    \end{cases}    
\end{equation}
where $R^{t-1}$ denotes the validation return in the last iteration. The selected arm will be favored if the policy's generalization ability improves. Thus, the informativeness ($\omega_1=0.2$) or novelty ($\omega_2=0.8$) will be focused on in different stages of training.

In the stage of parameter selection, we utilize the adaptive preference for parameters with both high informativeness and low-density scores. From the total sampled parameters $\{\xi_i\}_{i=1}^n$, we select a subset $\{\xi_i\}_{i=1}^m$ following:
\begin{equation}
     \{\xi_i\}_{i=1}^m = \min_{\{\xi_i\}_{i=1}^m \subset \{\xi_i\}_{i=1}^n} \sum_{i=1}^m (1 - \omega_{i^t}) I_i - \omega_{i^t} D_i
    \label{eq:selection}
\end{equation}
where $\omega_{i^t}$ is the sampled trade-off coefficient following Eq.~\ref{eq:prefrence_sampling} to control the preference between two indicators.

\subsection{Randomization Space Discretization}
\label{method:discrete}
We can simply apply the functions and evaluate the sampled parameters in the parameters selection stage. However, the main challenge is that the randomization space is continuous, in which case visited parameters rarely get sampled again in the following training process. One solution for the setting with uncountable tasks is to build a system parameter-conditioned policy which is similar to goal-conditional RL and hope the policy would generalize to unseen system parameters like prior works \cite{yu2017preparing, yu2018policy}. However, system parameters are different from the goals in the goal-conditioned RL \cite{andrychowicz2017hindsight, ghosh2019learning, chane2021goal} where goals belong to the same space as states and can provide implicit behavior signals for the policy. There is no evidence that the trained parameter-conditioned policy will generalize to unseen parameters by simply adjusting the input parameter values to the parameters in the target system. To mitigate the setting with uncountable tasks, we introduce a natural assumption:
\begin{assumption}
    \label{asp:PCH}
    (\textbf{Performance Continuity Hypothesis)} We denote the performance in expectation of a policy $\pi$ in a system with parameter $\xi$ as $J(\pi, \xi)$. Consider a policy $\pi_\theta(a|s)$, for any two system parameters $\xi_1, \xi_2 \in \Xi$, there exist $L > 0$ such that $|J(\pi_\theta, \xi_1) - J(\pi_\theta, \xi_2)| \leq L||\xi_1 - \xi_2||$.
\end{assumption}
The assumption states that the performance of any fixed policy under two different systems with close system parameters is close. Since Assumption~\ref{asp:PCH} implies the policy performance is similar in a relatively small range, we can approximate the randomization space $\Xi$ via discretization. Here we define a bin width $\nu$ and split each parameter range $(\xi_{low}^i, \xi_{high}^i)$ to several bins. By introducing discretization, we represent ISF and DF as two tables. Parameter evaluation is operated via simple look-up, and table values get updated naturally at each iteration. An alternative to discretization is to utilize two function approximators like neural networks to represent two functions and get updated via regression. However, we demonstrate that the discretization provides effective superiority considering the operation simplicity, which is empirically shown in Section~\ref{exp}.

\section{Experiments}
\label{exp}
In this section, we first describe our experimental setup and training details. Then, we discuss and answer the following key questions: (i) Does ADP produce more robust policies compared to baselines in the presence of system parameter errors; (ii) Can ADP handle domains with unmodeled disturbances? Here we investigate the system noise problems (e.g., sensor noise);  (iii) How do the sampled parameters change explicitly during training? (iv) How sensitive is ADP to design choices? (e.g., the utilization of the bandit)

\subsection{Environment Setup}
\label{exp:envset}
To evaluate the ability of ADP to improve policy robustness, we experiment with four continuous control tasks (Hopper, Walker, Ant, Halfcheetah) using Mujoco physics simulator~\cite{todorov2012mujoco}. We build different benchmarks where partial system parameters (e.g., torso mass, friction, etc.) are randomized in predefined ranges. We denote each benchmark as \textit{Environment-Parameters} (eg. \textit{Hopper - torso mass}). To evaluate the robustness of the policy, we define a training set, validation set, and test set with different parameter randomization ranges for each benchmark. The detailed setup of the benchmarks is presented in \ref{apd:env}.

\subsection{Implementation Details and Baseline Methods}
\label{exp:traindetail}
We adopt Proximal Policy Optimization (PPO)~\cite{schulman2017proximal} as our backbone RL algorithm. The policy is evaluated in the validation set periodically. We run each experiment with five seeds, and the average performance and the standard deviation are reported. Detailed implementation and hyperparameter setting are specified in \ref{apd:hyper}. 
To test the effectiveness of the active sampling strategy in ADP, we choose Uniform Domain Randomization (UDR)~\cite{peng2018sim} which uniformly samples tasks as the main baseline method. In addition, we compare ADP with two robust RL methods: RARL~\cite{pinto2017robust} and EPOpt~\cite{rajeswaran2016epopt}. Further details of baseline methods are shown in \ref{apd:baseline}.

\subsection{Robustness Evaluation}
\label{exp:robust}

\subsubsection{Robustness with Single Parameter Randomization}
\label{exp:robust:single}
\begin{figure}[t]
    \centering
    \includegraphics[width=\linewidth]{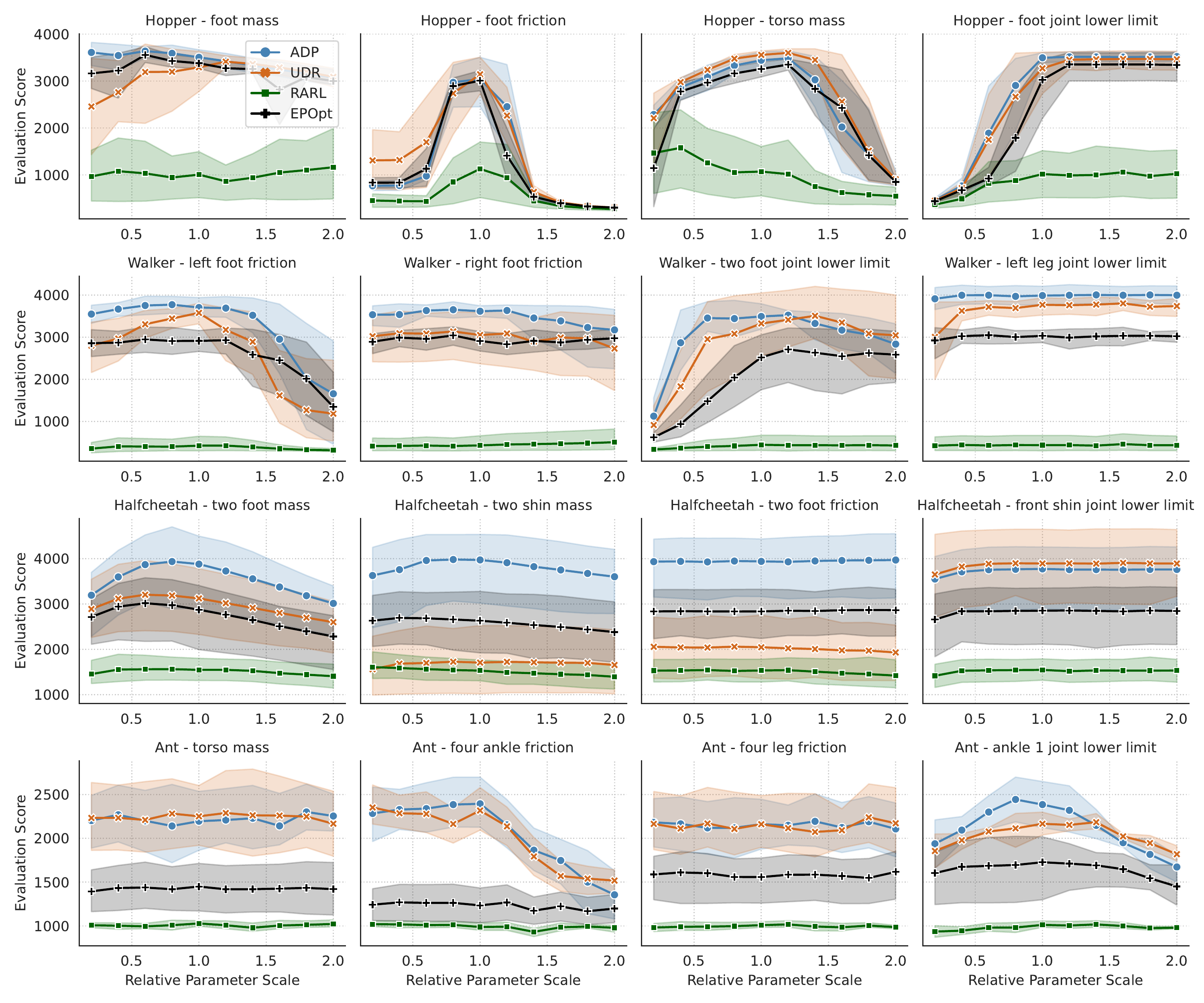}
    \caption{Robustness comparison under single system parameter randomization. After training the policy in the training set with randomization space $\text{default}\times[0.8,1.2]$, we evaluate the final policy in the test set with randomization space $\text{default}\times[0.2,2.0]$. We choose 10 discrete value $[0.2, 0.4,\dots, 2.0]$ from the parameter range. We evaluate each algorithm for 20 episodes. The shaded area represents the standard deviation across five random seeds.}
    \label{fig:single_coeff}
\end{figure}

To investigate whether ADP provides a robustness benefit for the policy, we start with experiments that randomized a single system parameter. We build 16 benchmarks by selecting 4 different system parameters for each environment. In each benchmark, we run all algorithms with 3 million timesteps in the training set. The default parameters are scaled by the values in $[0.8, 1.2]$ for the training set, which we denote as $\text{default}\times[0.8,1.2]$. To verify the policy performance in the test set, we discretize the parameter range of the test set whose randomization space is $\text{default}\times[0.2,2.0]$ and evaluate the policy for 20 episodes in each task.
As the results show in Figure~\ref{fig:single_coeff}, it is worth noting that there is no performance decline in some benchmarks even though the system parameter varies sharply like \textit{Halfcheetah - two shin mass} and \textit{Halfcheetah - two foot friction}. Here we denote these parameters as \textit{inactive parameters}, and we call parameters that influence the policy performance as \textit{active parameters}. We note that UDR performs poorly in benchmarks randomizing inactive parameters, while such system parameters are typically utilized in prior works for domain randomization~\cite{yu2017preparing,xie2021dynamics}. However, ADP significantly outperforms UDR in such benchmarks. We believe that the uniformly sampled tasks result in instable value targets for the TD-error calculation. The policy trained with an imprecise value function might behave poorly, as shown in the prior work~\cite{fujimoto2018addressing}. In contrast, ADP actively sampled tasks with relatively larger TD-errors, which produces stable training dynamics and superior policy performance.

\begin{figure}[t]
    \centering
    \includegraphics[width=\linewidth]{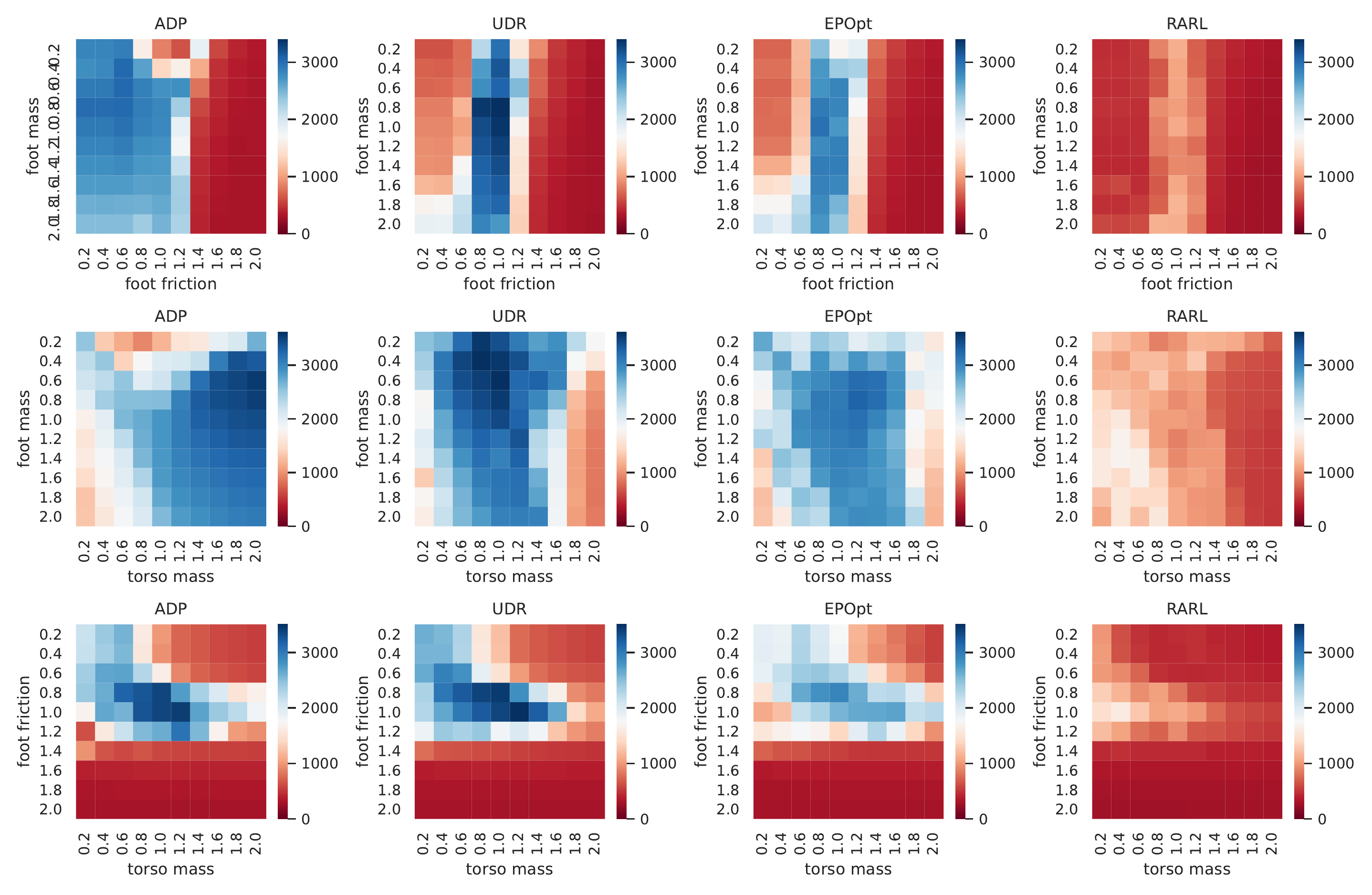}
    \caption{Robustness comparison under double system parameter randomization for Hopper. After training the policy in the train set with randomization space $\text{default}\times[0.8,1.2]$ for each parameter, we evaluate the final policy in the test set with randomization space $\text{default}\times[0.2,2.0]$ for 20 episodes. The rewards are averaged over five random seeds, and each average value is represented as a single grid. 
    }
    \label{fig:double_coeff_hopper_partial}
\end{figure}

Comparing ADP with UDR, we find that ADP shows superior performance in benchmarks of Walker and Halfcheetah, while similar performance is demonstrated in Hopper and Ant. Then we compare ADP with robust RL baselines. We notice that RARL obtains nearly the worst performance across all benchmarks, which could be explained by the conservative results introduced by the adversarial training. In addition, EPopt obtains similar results as UDR in some benchmarks. However, the performance of EPopt exceeds UDR in benchmarks with the inactivate parameters, which could derive from that the training preference on tasks at risk stables the training dynamic of the value function. 

\subsubsection{Robustness with Double Parameter Randomization}
\begin{figure}[htbp]
    \centering
    \includegraphics[width=\linewidth]{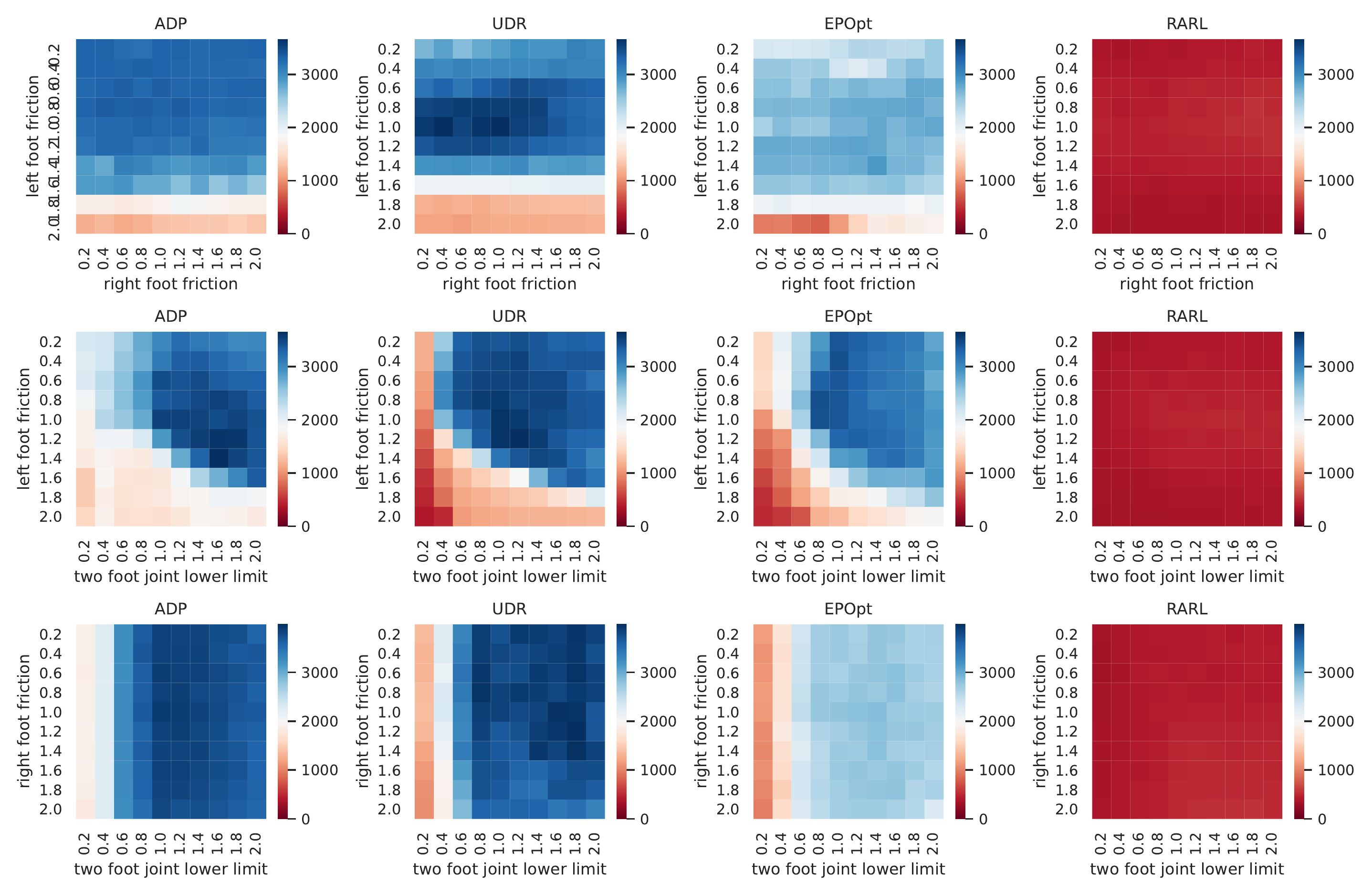}
    \caption{Robustness comparison under double system parameter randomization for Walker. After training the policy in the training set with randomization space $\text{default}\times[0.8,1.2]$ for each parameter, we evaluate the final policy in the test set with randomization space $\text{default}\times[0.2,2.0]$ for 20 episodes. The rewards are averaged over five random seeds, and each average value is represented as a single grid. 
    }
    \label{fig:double_coeff_walker_partial}
\end{figure}

In real-world systems, there could be different kinds of disturbances, such as external forces or internal system noise. It is critical to verify the policy robustness with multi-parameter randomization. However, as the number of randomization system parameters increases, the capacity of the implicit task set grows exponentially, while the interactions between parameters are uninterpretable. Here we conduct experiments that randomize double system parameters. We choose three parameters for Hopper and Walker, and combine every two parameters, which results in three benchmarks for each environment. We train each algorithm for 3 million timesteps in the train set, and the final policy is evaluated for 20 episodes in each task of the discretized test set. We demonstrate the results in Figure~\ref{fig:double_coeff_hopper_partial} and Figure~\ref{fig:double_coeff_walker_partial}.

As results show in Figure~\ref{fig:double_coeff_hopper_partial} and Figure~\ref{fig:double_coeff_walker_partial}, the policies trained by ADP demonstrate advanced generality and performance compared to all baseline methods when double parameters are randomized. When two system parameters are randomized, the increased benefit of the actively sampling strategy is expected due to the increased capacity of the test set intuitively.

\subsubsection{Robustness with Higher Dimensional Randomization}
\begin{table}
    \centering
    \caption{Robustness comparison with randomization over seven system parameters. We notice the policies trained via ADP outperform baseline approaches in three out of four environments.}
    \vspace{1em}
    \label{tab:seven_rand}
    \begin{tabular}{c|c|c|c}
        \toprule  
        ~ & EPOpt & UDR & ADP   \\  
        \midrule
        Hopper & \textbf{1038~$\pm$~693} & $804\pm464$ & $956\pm644$  \\  
        \midrule
        Walker & $1617\pm812$ & $1738\pm1033$ & \textbf{2155~$\pm$~1037} \\  \midrule
        Halfcheetah & $2167\pm411$ & $3048\pm653$ & \textbf{3093~$\pm$~649} \\
        \midrule
        Ant & $1307\pm250$ & $1733\pm345$ & \textbf{1793~$\pm$~299}     \\
        \bottomrule
    \end{tabular}
\end{table}
To further analyze the policy robustness in more challenging settings, we build benchmarks for all 4 environments where the number of randomization parameters is 7. The explicit randomized system parameters of each environment are demonstrated in~\ref{apd:env}. The final policy is evaluated in randomly sampled tasks from the test set. Since there are numerous tasks in the test set, we randomly sample 1000 tasks from the test set to evaluate each algorithm across five random seeds. 
Table~\ref{tab:seven_rand} depicts the policy robustness of each algorithm when seven system parameters are in randomization. As the results demonstrated, ADP shows improvement compared to UDR across all environments, and ADP outperforms EPOpt in most environments.

\subsection{Transfer to Domains with System Noise}
\label{exp:transfer}
Since prior works have emphasized that the noise of sensors~\cite{haojie2022reinforcement} or actuators~\cite{hwangbo2019learning} in the real-world robotics system makes the sim2real transfer difficult, we aim to investigate the robustness of our method in such general settings. Therefore, we evaluate our algorithm regarding observation noise and action noise in Hopper and Walker environments. 

\begin{figure}[htbp]
  \centering
  \includegraphics[width=\linewidth]{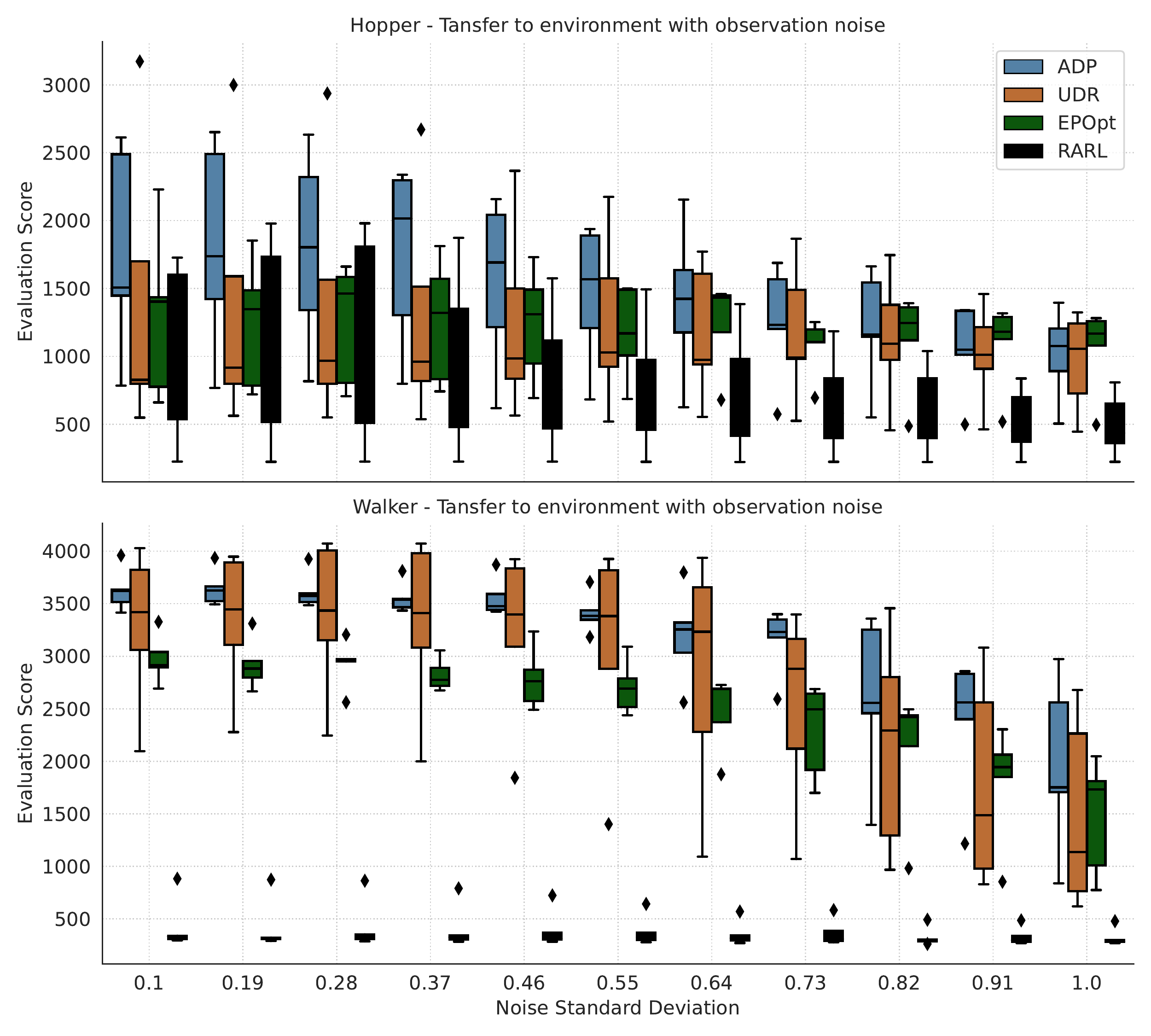}
  \caption{\textbf{Robustness to noisy observations}: ADP demonstrates superior robustness compared to all baseline methods in both environments.}
  \label{fig:transfer_obs}
\end{figure}

\textbf{Robustness to noisy observations.} We first evaluate algorithms in terms of the observation noise. In this experiment, we utilize Gaussian noise with different standard deviations $\sigma\in[0.1, 1.0]$ to perturb the velocity data in the observation. Before transferring to a domain with noisy observations, we train policies with randomization of seven system parameters shown in \ref{apd:env}. We report the zero-shot performance of the transferred policies for all algorithms, and the results are demonstrated in Figure~\ref{fig:transfer_obs}. In both environments, ADP shows better tolerance to the observation noise than all baseline methods. The robust policies yielded by ADP may prove useful in real-world environments where sensor measurements are inaccurate or corrupted.

\begin{figure}[htbp]
  \centering
  \includegraphics[width=\linewidth]{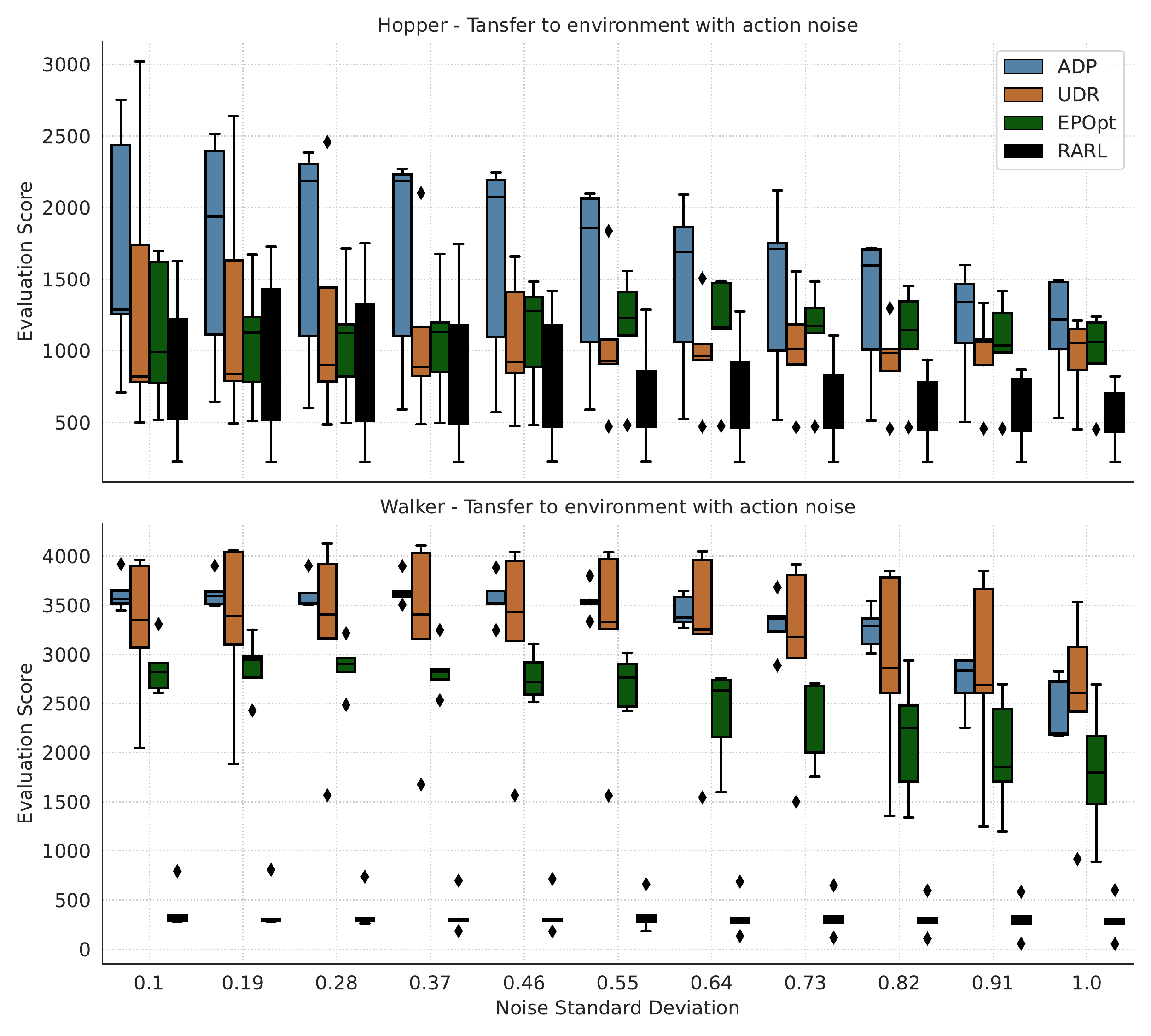}
  \caption{\textbf{Robustness to noisy actions}: ADP demonstrates better generalization to tasks with noise actions compared to all baseline methods in both environments.}
  \label{fig:transfer_action}
\end{figure}

\textbf{Robustness to noisy actions.} Then, we evaluate algorithms in terms of the action noise. We perturb the actions using Gaussian noise with different standard deviations $\sigma\in[0.1, 0.4]$. Same as before, we train policies with randomization of seven system parameters. We obtain each policy performance score averaged over 30 episodes. Figure~\ref{fig:transfer_action} shows that ADP is more robust to baseline methods in Walker. In Hopper, ADP outperforms robust RL baseline methods and shows more consistent performance across all runs comparing UDR.

\subsection{Evolution of the sampled parameters}

\begin{figure}[htbp]
  \centering
  \includegraphics[width=\linewidth]{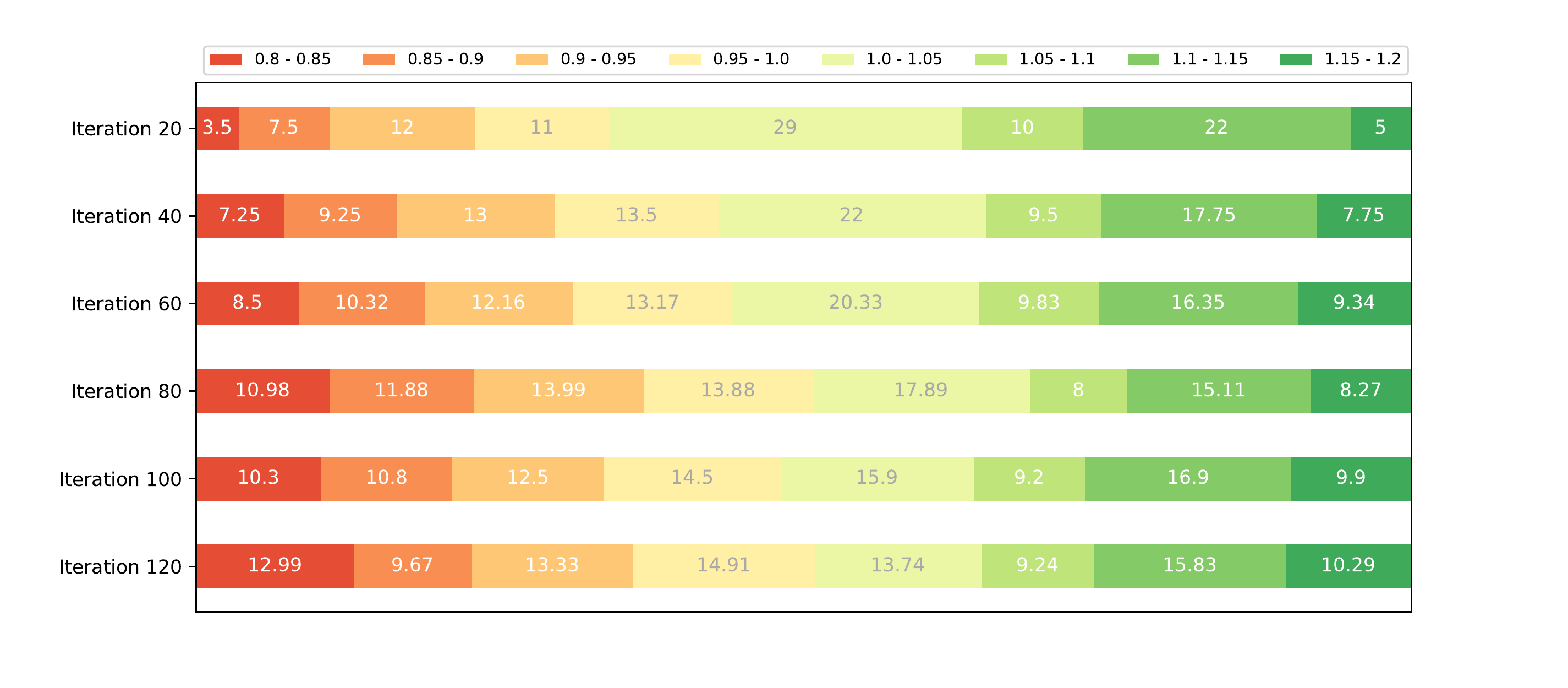}
  \caption{\textbf{Evolution of the percentages of sampled parameters during training in \textit{Hopper - foot mass}}: The different colors present parameters in different intervals, and the numbers in the color bars represent the percentage of times the parameters in the corresponding interval have been sampled. We can observe that the systems with a light foot~(foot mass $\in (0.8,0.5]$) or a heavy foot~(foot mass $\in (1.15,1.2]$) are less sampled in the early training stage and are increasingly sampled as the capability of the agent improves.}
  \label{fig:coeff_evolve}
\end{figure}

Here we claim that ADP produces some meaningful sampling order for the candidate parameters, which enhances the generalization ability of the agent. To prove the hypothesis, we visualize the percentages of sampled parameters in different intervals in \textit{Hopper - foot mass} during training. As the result shown in Figure~\ref{fig:coeff_evolve}, the systems with proper foot mass~(foot mass $\in(0.95,1.05]$) are sampled more frequently at the early training stage, while the systems with a light foot~(foot mass $\in (0.8,0.5]$) or a heavy foot~(foot mass $\in (1.15,1.2]$) are increasingly sampled as the training progresses. It is worth noting that hopping without falling down with a light foot or heavy foot is difficult for Hopper. Thus, we can observe that ADP generates an implicit curriculum during training which has been proved to benefit the generalization ability~\cite{fang2019curriculum,jiang2021prioritized}.

\subsection{Ablations on ADP}
\label{exp:ablation}
We run ablation experiments to understand the effects of various design choices in implementing ADP. Specially, we analyze the effect of (i) bin width for the randomization space discretization and (ii) preference correction via the bandit. 

\begin{wrapfigure}{r}{0.4\textwidth} 
    \centering
    \vspace{-5mm}
    \includegraphics[width=0.4\textwidth]{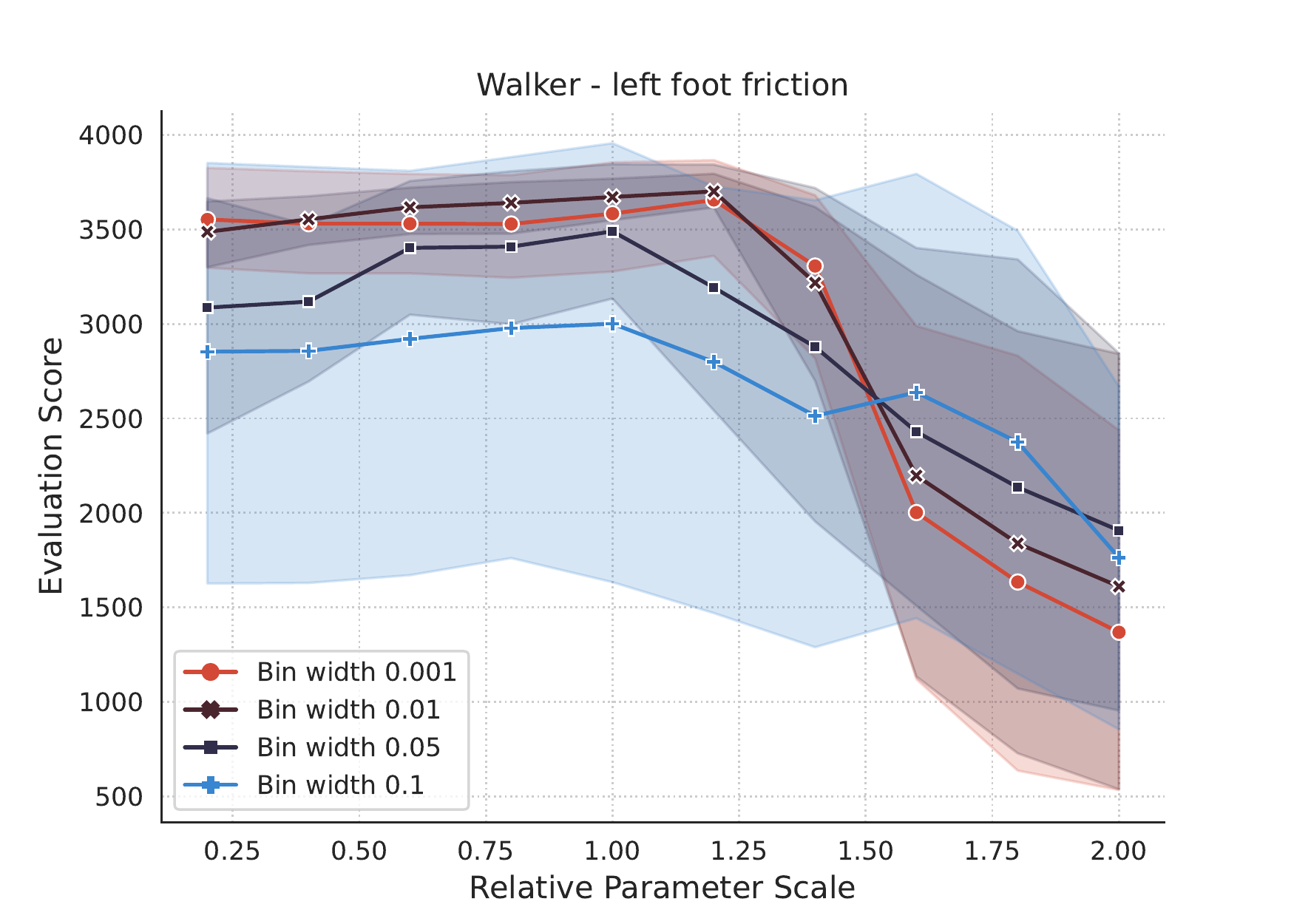}
    \captionof{figure}{Investigation for the effect of various bin widths. 
    }
    \label{fig:abalation_bin}
\end{wrapfigure}
\subsubsection{How important is the choice of bin width?}
We discretize the randomization space based on the performance continuity assumption mentioned in Section~\ref{method:discrete}. However, it is unclear how the utilized bin width influences the algorithm's performance. Intuitively, performance degradation might happen when utilizing overly wide or narrow bins. In terms of the parameter scale range $[0.8, 1.2]$, we compare the effect of various bin widths $\{0.001, 0.01, 0.05, 0.1\}$ in the \textit{Walker - left foot friction} benchmark. As the results shown in Figure~\ref{fig:abalation_bin}, we notice overly large bin width like $0.1$ results in worse performance compared to others. In contrast, the performance is not influenced by utilizing the smallest bin width $0.001$, and we believe that the performance continuity assumption proposed in Section \ref{method:discrete} can support the results.

\subsubsection{How important is the adaptive preference correction via bandit?}
\begin{figure}[htbp]
    \centering
    \subfigure[]{\includegraphics[width=0.48\textwidth]{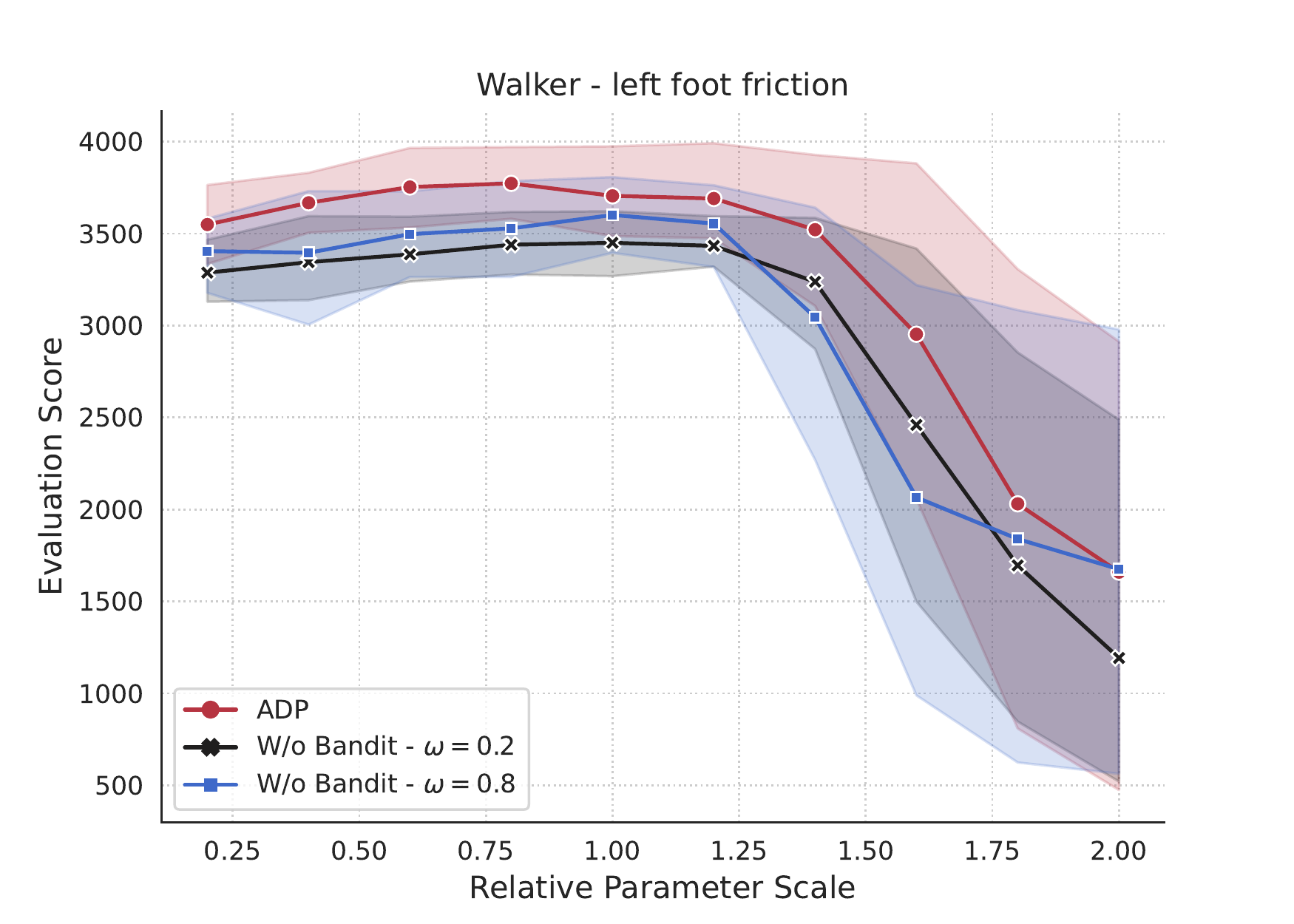}} 
    \subfigure[]{\includegraphics[width=0.48\textwidth]{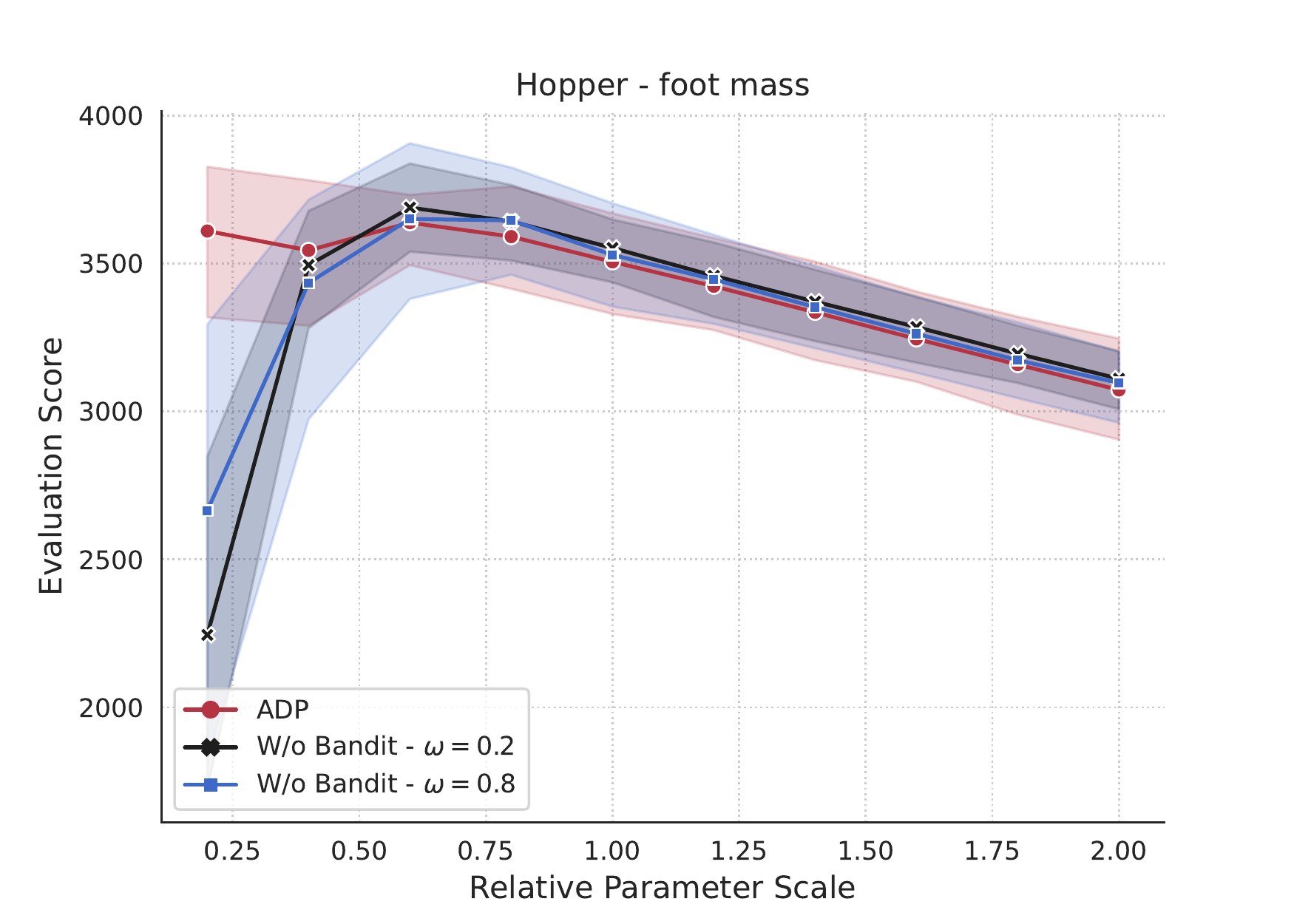}}
    \captionof{figure}{We investigate the effect of the adaptive preference correction via the bandit in (a)~\textit{Walker - left foot friction} and (b)~\textit{Hopper - foot mass}. The results are run across 5 random seeds. The results indicate that introducing the bandit improves the policy robustness compared to the variants with fixed trade-off coefficients.}
    \label{fig:abalation_bandit}
    
\end{figure}

To investigate if the adaptive preference correction improves the robustness of the final policy, we implement two variants of ADP with fixed $\alpha$ trade-off, including ${\alpha=0.2, \alpha=0.8}$. We compare the policy robustness of the three variants in the \textit{Walker-left foot friction} and \textit{Hopper - foot mass}. The results shown in Figure~\ref{fig:abalation_bandit} demonstrate that adaptive preference correction contributes to the performance of ADP. Especially, both variants with fixed trade-off coefficients fail in Hopper with a light foot~(foot mass scale $=0.2$), while ADP shows robust performance in the challenging environment. Thus, the adaptive preference correction via bandit is beneficial to the generalization of ADP.

\section{Conclusion and Future Work}
\label{conclusion}
In this work, we present Active Dynamics Preference~(ADP) to improve Randomization~(DR) through active task sampling. By introducing the performance continuity hypothesis heuristically, we transfer the infinite task setting to a countable task setting, and informativeness and novelty of system parameters are utilized to guide the task sampling. Extensive experiments reveal the existence of some inactive system parameters as the cause of the over-conservative policy problem of DR, and we demonstrate that ADP outperforms several baseline methods in terms of various model discrepancies. 

For future research, we believe ADP can be extended to the sim2real of robotics as a more powerful alternative to DR. Furthermore, providing theoretical evidence for the generalization ability might be promising to interpret the empirical results of ADP and other works which incorporate active task sampling technique~\cite{florensa2018automatic, portelas2020teacher, fang2019curriculum}.

\bibliographystyle{IEEEtran}
\bibliography{ref}

\begin{thebibliography}{10}
\providecommand{\url}[1]{#1}
\csname url@samestyle\endcsname
\providecommand{\newblock}{\relax}
\providecommand{\bibinfo}[2]{#2}
\providecommand{\BIBentrySTDinterwordspacing}{\spaceskip=0pt\relax}
\providecommand{\BIBentryALTinterwordstretchfactor}{4}
\providecommand{\BIBentryALTinterwordspacing}{\spaceskip=\fontdimen2\font plus
\BIBentryALTinterwordstretchfactor\fontdimen3\font minus
  \fontdimen4\font\relax}
\providecommand{\BIBforeignlanguage}[2]{{%
\expandafter\ifx\csname l@#1\endcsname\relax
\typeout{** WARNING: IEEEtran.bst: No hyphenation pattern has been}%
\typeout{** loaded for the language `#1'. Using the pattern for}%
\typeout{** the default language instead.}%
\else
\language=\csname l@#1\endcsname
\fi
#2}}
\providecommand{\BIBdecl}{\relax}
\BIBdecl

\bibitem{ecoffet2021first}
A.~Ecoffet, J.~Huizinga, J.~Lehman, K.~O. Stanley, and J.~Clune, ``First
  return, then explore,'' \emph{Nature}, vol. 590, no. 7847, pp. 580--586,
  2021.

\bibitem{jaderberg2019human}
M.~Jaderberg, W.~M. Czarnecki, I.~Dunning, L.~Marris, G.~Lever, A.~G.
  Castaneda, C.~Beattie, N.~C. Rabinowitz, A.~S. Morcos, A.~Ruderman
  \emph{et~al.}, ``Human-level performance in 3d multiplayer games with
  population-based reinforcement learning,'' \emph{Science}, vol. 364, no.
  6443, pp. 859--865, 2019.

\bibitem{levine2018learning}
S.~Levine, P.~Pastor, A.~Krizhevsky, J.~Ibarz, and D.~Quillen, ``Learning
  hand-eye coordination for robotic grasping with deep learning and large-scale
  data collection,'' \emph{The International journal of robotics research},
  vol.~37, no. 4-5, pp. 421--436, 2018.

\bibitem{andrychowicz2020learning}
O.~M. Andrychowicz, B.~Baker, M.~Chociej, R.~Jozefowicz, B.~McGrew,
  J.~Pachocki, A.~Petron, M.~Plappert, G.~Powell, A.~Ray \emph{et~al.},
  ``Learning dexterous in-hand manipulation,'' \emph{The International Journal
  of Robotics Research}, vol.~39, no.~1, pp. 3--20, 2020.

\bibitem{peng2020learning}
X.~B. Peng, E.~Coumans, T.~Zhang, T.-W. Lee, J.~Tan, and S.~Levine, ``Learning
  agile robotic locomotion skills by imitating animals,'' \emph{arXiv preprint
  arXiv:2004.00784}, 2020.

\bibitem{hwangbo2019learning}
J.~Hwangbo, J.~Lee, A.~Dosovitskiy, D.~Bellicoso, V.~Tsounis, V.~Koltun, and
  M.~Hutter, ``Learning agile and dynamic motor skills for legged robots,''
  \emph{Science Robotics}, vol.~4, no.~26, p. eaau5872, 2019.

\bibitem{yu2017preparing}
W.~Yu, J.~Tan, C.~K. Liu, and G.~Turk, ``Preparing for the unknown: Learning a
  universal policy with online system identification,'' \emph{arXiv preprint
  arXiv:1702.02453}, 2017.

\bibitem{peng2018sim}
X.~B. Peng, M.~Andrychowicz, W.~Zaremba, and P.~Abbeel, ``Sim-to-real transfer
  of robotic control with dynamics randomization,'' in \emph{2018 IEEE
  international conference on robotics and automation (ICRA)}.\hskip 1em plus
  0.5em minus 0.4em\relax IEEE, 2018, pp. 3803--3810.

\bibitem{ross2011reduction}
S.~Ross, G.~Gordon, and D.~Bagnell, ``A reduction of imitation learning and
  structured prediction to no-regret online learning,'' in \emph{Proceedings of
  the fourteenth international conference on artificial intelligence and
  statistics}.\hskip 1em plus 0.5em minus 0.4em\relax JMLR Workshop and
  Conference Proceedings, 2011, pp. 627--635.

\bibitem{tobin2017domain}
J.~Tobin, R.~Fong, A.~Ray, J.~Schneider, W.~Zaremba, and P.~Abbeel, ``Domain
  randomization for transferring deep neural networks from simulation to the
  real world,'' in \emph{2017 IEEE/RSJ international conference on intelligent
  robots and systems (IROS)}.\hskip 1em plus 0.5em minus 0.4em\relax IEEE,
  2017, pp. 23--30.

\bibitem{xie2021dynamics}
Z.~Xie, X.~Da, M.~van~de Panne, B.~Babich, and A.~Garg, ``Dynamics
  randomization revisited: A case study for quadrupedal locomotion,'' in
  \emph{2021 IEEE International Conference on Robotics and Automation
  (ICRA)}.\hskip 1em plus 0.5em minus 0.4em\relax IEEE, 2021, pp. 4955--4961.

\bibitem{mehta2020active}
B.~Mehta, M.~Diaz, F.~Golemo, C.~J. Pal, and L.~Paull, ``Active domain
  randomization,'' in \emph{Conference on Robot Learning}.\hskip 1em plus 0.5em
  minus 0.4em\relax PMLR, 2020, pp. 1162--1176.

\bibitem{toneva2018empirical}
M.~Toneva, A.~Sordoni, R.~T.~d. Combes, A.~Trischler, Y.~Bengio, and G.~J.
  Gordon, ``An empirical study of example forgetting during deep neural network
  learning,'' \emph{arXiv preprint arXiv:1812.05159}, 2018.

\bibitem{florensa2018automatic}
C.~Florensa, D.~Held, X.~Geng, and P.~Abbeel, ``Automatic goal generation for
  reinforcement learning agents,'' in \emph{International conference on machine
  learning}.\hskip 1em plus 0.5em minus 0.4em\relax PMLR, 2018, pp. 1515--1528.

\bibitem{zhang2020automatic}
Y.~Zhang, P.~Abbeel, and L.~Pinto, ``Automatic curriculum learning through
  value disagreement,'' \emph{Advances in Neural Information Processing
  Systems}, vol.~33, pp. 7648--7659, 2020.

\bibitem{jiang2021prioritized}
M.~Jiang, E.~Grefenstette, and T.~Rockt{\"a}schel, ``Prioritized level
  replay,'' in \emph{International Conference on Machine Learning}.\hskip 1em
  plus 0.5em minus 0.4em\relax PMLR, 2021, pp. 4940--4950.

\bibitem{mozian2020learning}
M.~Mozian, J.~C.~G. Higuera, D.~Meger, and G.~Dudek, ``Learning domain
  randomization distributions for training robust locomotion policies,'' in
  \emph{2020 IEEE/RSJ International Conference on Intelligent Robots and
  Systems (IROS)}.\hskip 1em plus 0.5em minus 0.4em\relax IEEE, 2020, pp.
  6112--6117.

\bibitem{rajeswaran2016epopt}
A.~Rajeswaran, S.~Ghotra, B.~Ravindran, and S.~Levine, ``Epopt: Learning robust
  neural network policies using model ensembles,'' \emph{arXiv preprint
  arXiv:1610.01283}, 2016.

\bibitem{ramos2019bayessim}
F.~Ramos, R.~C. Possas, and D.~Fox, ``Bayessim: adaptive domain randomization
  via probabilistic inference for robotics simulators,'' \emph{arXiv preprint
  arXiv:1906.01728}, 2019.

\bibitem{muratore2021data}
F.~Muratore, C.~Eilers, M.~Gienger, and J.~Peters, ``Data-efficient domain
  randomization with bayesian optimization,'' \emph{IEEE Robotics and
  Automation Letters}, vol.~6, no.~2, pp. 911--918, 2021.

\bibitem{liu2017stein}
Y.~Liu, P.~Ramachandran, Q.~Liu, and J.~Peng, ``Stein variational policy
  gradient,'' in \emph{33rd Conference on Uncertainty in Artificial
  Intelligence, UAI 2017}, 2017.

\bibitem{eysenbach2018diversity}
B.~Eysenbach, A.~Gupta, J.~Ibarz, and S.~Levine, ``Diversity is all you need:
  Learning skills without a reward function,'' \emph{arXiv preprint
  arXiv:1802.06070}, 2018.

\bibitem{tamar2015optimizing}
A.~Tamar, Y.~Glassner, and S.~Mannor, ``Optimizing the cvar via sampling,'' in
  \emph{Twenty-Ninth AAAI Conference on Artificial Intelligence}, 2015.

\bibitem{snoek2012practical}
J.~Snoek, H.~Larochelle, and R.~P. Adams, ``Practical bayesian optimization of
  machine learning algorithms,'' \emph{Advances in neural information
  processing systems}, vol.~25, 2012.

\bibitem{pinto2017robust}
L.~Pinto, J.~Davidson, R.~Sukthankar, and A.~Gupta, ``Robust adversarial
  reinforcement learning,'' in \emph{International Conference on Machine
  Learning}.\hskip 1em plus 0.5em minus 0.4em\relax PMLR, 2017, pp. 2817--2826.

\bibitem{tessler2019action}
C.~Tessler, Y.~Efroni, and S.~Mannor, ``Action robust reinforcement learning
  and applications in continuous control,'' in \emph{International Conference
  on Machine Learning}.\hskip 1em plus 0.5em minus 0.4em\relax PMLR, 2019, pp.
  6215--6224.

\bibitem{kamalaruban2020robust}
P.~Kamalaruban, Y.-T. Huang, Y.-P. Hsieh, P.~Rolland, C.~Shi, and V.~Cevher,
  ``Robust reinforcement learning via adversarial training with langevin
  dynamics,'' \emph{Advances in Neural Information Processing Systems},
  vol.~33, pp. 8127--8138, 2020.

\bibitem{vinitsky2020robust}
E.~Vinitsky, Y.~Du, K.~Parvate, K.~Jang, P.~Abbeel, and A.~Bayen, ``Robust
  reinforcement learning using adversarial populations,'' \emph{arXiv preprint
  arXiv:2008.01825}, 2020.

\bibitem{jiang2021monotonic}
Y.~Jiang, C.~Li, W.~Dai, J.~Zou, and H.~Xiong, ``Monotonic robust policy
  optimization with model discrepancy,'' in \emph{International Conference on
  Machine Learning}.\hskip 1em plus 0.5em minus 0.4em\relax PMLR, 2021, pp.
  4951--4960.

\bibitem{wang2021online}
Y.~Wang and S.~Zou, ``Online robust reinforcement learning with model
  uncertainty,'' \emph{Advances in Neural Information Processing Systems},
  vol.~34, 2021.

\bibitem{eysenbach2021robust}
B.~Eysenbach, R.~R. Salakhutdinov, and S.~Levine, ``Robust predictable
  control,'' \emph{Advances in Neural Information Processing Systems}, vol.~34,
  2021.

\bibitem{baranes2009r}
A.~Baranes and P.-Y. Oudeyer, ``R-iac: Robust intrinsically motivated
  exploration and active learning,'' \emph{IEEE Transactions on Autonomous
  Mental Development}, vol.~1, no.~3, pp. 155--169, 2009.

\bibitem{florensa2017reverse}
C.~Florensa, D.~Held, M.~Wulfmeier, M.~Zhang, and P.~Abbeel, ``Reverse
  curriculum generation for reinforcement learning,'' in \emph{Conference on
  robot learning}.\hskip 1em plus 0.5em minus 0.4em\relax PMLR, 2017, pp.
  482--495.

\bibitem{ivanovic2019barc}
B.~Ivanovic, J.~Harrison, A.~Sharma, M.~Chen, and M.~Pavone, ``Barc: Backward
  reachability curriculum for robotic reinforcement learning,'' in \emph{2019
  International Conference on Robotics and Automation (ICRA)}.\hskip 1em plus
  0.5em minus 0.4em\relax IEEE, 2019, pp. 15--21.

\bibitem{matiisen2019teacher}
T.~Matiisen, A.~Oliver, T.~Cohen, and J.~Schulman, ``Teacher--student
  curriculum learning,'' \emph{IEEE transactions on neural networks and
  learning systems}, vol.~31, no.~9, pp. 3732--3740, 2019.

\bibitem{portelas2020teacher}
R.~Portelas, C.~Colas, K.~Hofmann, and P.-Y. Oudeyer, ``Teacher algorithms for
  curriculum learning of deep rl in continuously parameterized environments,''
  in \emph{Conference on Robot Learning}.\hskip 1em plus 0.5em minus
  0.4em\relax PMLR, 2020, pp. 835--853.

\bibitem{baker2019emergent}
B.~Baker, I.~Kanitscheider, T.~Markov, Y.~Wu, G.~Powell, B.~McGrew, and
  I.~Mordatch, ``Emergent tool use from multi-agent autocurricula,''
  \emph{arXiv preprint arXiv:1909.07528}, 2019.

\bibitem{andrychowicz2017hindsight}
M.~Andrychowicz, F.~Wolski, A.~Ray, J.~Schneider, R.~Fong, P.~Welinder,
  B.~McGrew, J.~Tobin, O.~Pieter~Abbeel, and W.~Zaremba, ``Hindsight experience
  replay,'' \emph{Advances in neural information processing systems}, vol.~30,
  2017.

\bibitem{racaniere2019automated}
S.~Racaniere, A.~K. Lampinen, A.~Santoro, D.~P. Reichert, V.~Firoiu, and T.~P.
  Lillicrap, ``Automated curricula through setter-solver interactions,''
  \emph{arXiv preprint arXiv:1909.12892}, 2019.

\bibitem{colas2019curious}
C.~Colas, P.~Fournier, M.~Chetouani, O.~Sigaud, and P.-Y. Oudeyer, ``Curious:
  intrinsically motivated modular multi-goal reinforcement learning,'' in
  \emph{International conference on machine learning}.\hskip 1em plus 0.5em
  minus 0.4em\relax PMLR, 2019, pp. 1331--1340.

\bibitem{fang2019curriculum}
M.~Fang, T.~Zhou, Y.~Du, L.~Han, and Z.~Zhang, ``Curriculum-guided hindsight
  experience replay,'' \emph{Advances in neural information processing
  systems}, vol.~32, 2019.

\bibitem{zhao2019curiosity}
R.~Zhao and V.~Tresp, ``Curiosity-driven experience prioritization via density
  estimation,'' \emph{arXiv preprint arXiv:1902.08039}, 2019.

\bibitem{jiang2021replay}
M.~Jiang, M.~Dennis, J.~Parker-Holder, J.~Foerster, E.~Grefenstette, and
  T.~Rockt{\"a}schel, ``Replay-guided adversarial environment design,''
  \emph{Advances in Neural Information Processing Systems}, vol.~34, pp.
  1884--1897, 2021.

\bibitem{cobbe2020leveraging}
K.~Cobbe, C.~Hesse, J.~Hilton, and J.~Schulman, ``Leveraging procedural
  generation to benchmark reinforcement learning,'' in \emph{International
  conference on machine learning}.\hskip 1em plus 0.5em minus 0.4em\relax PMLR,
  2020, pp. 2048--2056.

\bibitem{kumar2021rma}
A.~Kumar, Z.~Fu, D.~Pathak, and J.~Malik, ``Rma: Rapid motor adaptation for
  legged robots,'' \emph{arXiv preprint arXiv:2107.04034}, 2021.

\bibitem{du2021auto}
Y.~Du, O.~Watkins, T.~Darrell, P.~Abbeel, and D.~Pathak, ``Auto-tuned
  sim-to-real transfer,'' in \emph{2021 IEEE International Conference on
  Robotics and Automation (ICRA)}.\hskip 1em plus 0.5em minus 0.4em\relax IEEE,
  2021, pp. 1290--1296.

\bibitem{puterman1990markov}
M.~L. Puterman, ``Markov decision processes,'' \emph{Handbooks in operations
  research and management science}, vol.~2, pp. 331--434, 1990.

\bibitem{schulman2015high}
J.~Schulman, P.~Moritz, S.~Levine, M.~Jordan, and P.~Abbeel, ``High-dimensional
  continuous control using generalized advantage estimation,'' \emph{arXiv
  preprint arXiv:1506.02438}, 2015.

\bibitem{schulman2015trust}
J.~Schulman, S.~Levine, P.~Abbeel, M.~Jordan, and P.~Moritz, ``Trust region
  policy optimization,'' in \emph{International conference on machine
  learning}.\hskip 1em plus 0.5em minus 0.4em\relax PMLR, 2015, pp. 1889--1897.

\bibitem{schulman2017proximal}
J.~Schulman, F.~Wolski, P.~Dhariwal, A.~Radford, and O.~Klimov, ``Proximal
  policy optimization algorithms,'' \emph{arXiv preprint arXiv:1707.06347},
  2017.

\bibitem{bellemare2016unifying}
M.~Bellemare, S.~Srinivasan, G.~Ostrovski, T.~Schaul, D.~Saxton, and R.~Munos,
  ``Unifying count-based exploration and intrinsic motivation,'' \emph{Advances
  in neural information processing systems}, vol.~29, 2016.

\bibitem{agrawal2012analysis}
S.~Agrawal and N.~Goyal, ``Analysis of thompson sampling for the multi-armed
  bandit problem,'' in \emph{Conference on learning theory}.\hskip 1em plus
  0.5em minus 0.4em\relax JMLR Workshop and Conference Proceedings, 2012, pp.
  39--1.

\bibitem{russo2016information}
D.~Russo and B.~Van~Roy, ``An information-theoretic analysis of thompson
  sampling,'' \emph{The Journal of Machine Learning Research}, vol.~17, no.~1,
  pp. 2442--2471, 2016.

\bibitem{yu2018policy}
W.~Yu, C.~K. Liu, and G.~Turk, ``Policy transfer with strategy optimization,''
  \emph{arXiv preprint arXiv:1810.05751}, 2018.

\bibitem{ghosh2019learning}
D.~Ghosh, A.~Gupta, A.~Reddy, J.~Fu, C.~Devin, B.~Eysenbach, and S.~Levine,
  ``Learning to reach goals via iterated supervised learning,'' \emph{arXiv
  preprint arXiv:1912.06088}, 2019.

\bibitem{chane2021goal}
E.~Chane-Sane, C.~Schmid, and I.~Laptev, ``Goal-conditioned reinforcement
  learning with imagined subgoals,'' in \emph{International Conference on
  Machine Learning}.\hskip 1em plus 0.5em minus 0.4em\relax PMLR, 2021, pp.
  1430--1440.

\bibitem{todorov2012mujoco}
E.~Todorov, T.~Erez, and Y.~Tassa, ``Mujoco: A physics engine for model-based
  control,'' in \emph{2012 IEEE/RSJ international conference on intelligent
  robots and systems}.\hskip 1em plus 0.5em minus 0.4em\relax IEEE, 2012, pp.
  5026--5033.

\bibitem{fujimoto2018addressing}
S.~Fujimoto, H.~Hoof, and D.~Meger, ``Addressing function approximation error
  in actor-critic methods,'' in \emph{International conference on machine
  learning}.\hskip 1em plus 0.5em minus 0.4em\relax PMLR, 2018, pp. 1587--1596.

\bibitem{haojie2022reinforcement}
S.~Haojie, B.~Zhou, H.~Zeng, F.~Wang, Y.~Dong, J.~Li, K.~Wang, H.~Tian, and
  M.~Q.-H. Meng, ``Reinforcement learning with evolutionary trajectory
  generator: A general approach for quadrupedal locomotion,'' \emph{IEEE
  Robotics and Automation Letters}, 2022.

\bibitem{brockman2016openai}
G.~Brockman, V.~Cheung, L.~Pettersson, J.~Schneider, J.~Schulman, J.~Tang, and
  W.~Zaremba, ``Openai gym,'' \emph{arXiv preprint arXiv:1606.01540}, 2016.

\bibitem{moritz2018ray}
P.~Moritz, R.~Nishihara, S.~Wang, A.~Tumanov, R.~Liaw, E.~Liang, M.~Elibol,
  Z.~Yang, W.~Paul, M.~I. Jordan \emph{et~al.}, ``Ray: A distributed framework
  for emerging $\{$AI$\}$ applications,'' in \emph{13th USENIX Symposium on
  Operating Systems Design and Implementation (OSDI 18)}, 2018, pp. 561--577.

\end{thebibliography}

\clearpage

\section{Environment Details}
\label{apd:env}

\subsection{Basic Environment Information}
For the four continuous control tasks, we follow the original setting from the OpenAI Gym~\cite{brockman2016openai}. The observation space, action space, and reward function of each environment are shown in Table~\ref{tab:basic_env}.

\begin{table*}[htb]
    \centering
    \caption{Basic information of environments}
    \resizebox{\linewidth}{!}{
        \begin{tabular}{c|c|c|c}
            
            Environment & Observation Space Dim & Action Space Dim & Reward Function \\
            \hline\hline
            \rule{0pt}{10pt}
            Hopper & 11 & 3 & $v_x - 0.001 * ||a||_2^2 + 1.0$\\
            \rule{0pt}{10pt}
            Walker & 17 & 6 & $v_x - 0.001 * ||a||_2^2 + 1.0$\\
            \rule{0pt}{10pt}
            Halfcheetah & 17 & 6 & $v_x - 0.1 * ||a||_2^2$ \\
            \rule{0pt}{10pt}
            Ant & 111 & 8 & $vel_x - 0.5 * ||a||_2^2 - 0.0005 * ||ext \_force||_2^2 + 1.0$
            
        \end{tabular}
    }
    \label{tab:basic_env}
\end{table*}

\subsection{Detailed Setup}
To evaluate the robustness of policies in held-out tasks, we create a validation set and test set for each benchmark by adjusting the parameter randomization space. The validation set is utilized for bandit update as noted in Section~\ref{method:bandit} and the best model selection during training. The test set is utilized to evaluate policy robustness.

\begin{figure*}[htb]
    \centering
    \includegraphics[width=\linewidth]{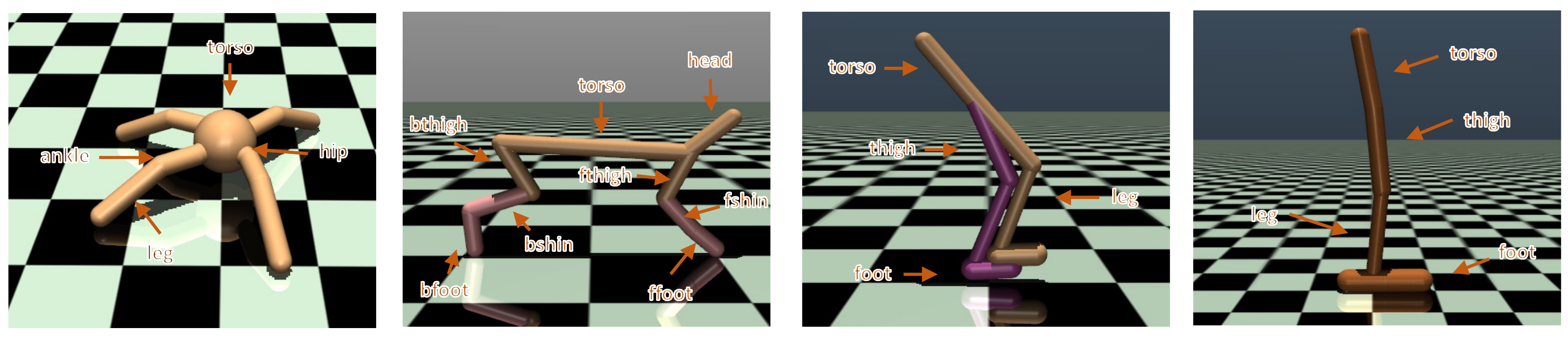}
    \caption{The body segments of Ant, Halfcheetah, Walker, Hopper.}
    \label{fig:body}
\end{figure*}

Below we list the system parameters randomization details used in our experiments. The randomization details for all parameters of each environment is shown in Tables~\ref{tab:random_hopper},~\ref{tab:random_walker},~\ref{tab:random_ant},~\ref{tab:random_half}. The body names in all environments are shown in Figure~\ref{fig:body}.

\begin{table*}[htb]
    \centering
    \caption{Parameter Randomization details for Hopper}
    \resizebox{\linewidth}{!}{
    \begin{tabular}{c|c|c|c|c}
        System Parameter & Default Value & Training Set & Validation Set & Test Set \\
        \hline\hline
        \rule{0pt}{8pt}
        foot mass & 5.089~kg &  & & \\
        \rule{0pt}{8pt}
        leg mass & 2.714~kg & & & \\
        \rule{0pt}{8pt}
        torso mass & 3.534~kg & & & \\
        \rule{0pt}{8pt}
        foot friction & 2.0 & $\text{default} \times [0.8, 1.2]$ & $\text{default} \times [0.5, 1.5]$ & $\text{default} \times [0.2, 2.0]$ \\
        \rule{0pt}{8pt}
        damping & 1.0 & & & \\
        \rule{0pt}{8pt}
        foot joint lower limit & -0.785 & & & \\
        \rule{0pt}{8pt}
        leg joint lower limit & -2.618 & & & 
    \end{tabular}
    }
    \label{tab:random_hopper}
\end{table*}

\begin{table*}[htb]
    \centering
    \caption{Parameter Randomization details for Walker}
    \resizebox{\linewidth}{!}{
    \begin{tabular}{c|c|c|c|c}
        System Parameter & Default Value & Training Set & Validation Set & Test Set \\
        \hline\hline
        \rule{0pt}{8pt}
        torso mass & 3.534~kg & & & \\
        \rule{0pt}{8pt}
        two foot mass & 3.534~kg & & & \\
        \rule{0pt}{8pt}
        left foot friction & 1.9 & & & \\
        \rule{0pt}{8pt}
        right foot friction & 0.9 & $\text{default} \times [0.8, 1.2]$ & $\text{default} \times [0.5, 1.5]$ & $\text{default} \times [0.2, 2.0]$ \\
        \rule{0pt}{8pt}
        two foot joint lower limit & -0.785 & & & \\
        \rule{0pt}{8pt}
        left leg joint lower limit & -2.618 & & & \\
        \rule{0pt}{8pt}
        right leg joint lower limit & -2.618 & & &
    \end{tabular}
    }
    \label{tab:random_walker}
\end{table*}

\begin{table*}[htb]
    \centering
    \caption{Parameter Randomization details for Ant}
    \resizebox{\linewidth}{!}{
    \begin{tabular}{c|c|c|c|c}
        System Parameter & Default Value & Training Set & Validation Set & Test Set \\
        \hline\hline
        \rule{0pt}{8pt}
        torso mass & 0.327~kg & & & \\
        \rule{0pt}{8pt}
        four leg mass & 0.036~kg & & & \\
        \rule{0pt}{8pt}
        four ankle friction & 1.0 & & & \\
        \rule{0pt}{8pt}
        four leg friction & 1.0 & $\text{default} \times [0.8, 1.2]$ & $\text{default} \times [0.5, 1.5]$ & $\text{default} \times [0.2, 2.0]$ \\
        \rule{0pt}{8pt}
        damping & 1.0 & & & \\
        \rule{0pt}{8pt}
        hip 1 joint lower limit & -0.524 & & & \\
        \rule{0pt}{8pt}
        ankle 1 joint lower limit & 0.524 & & &
    \end{tabular}
    }
    \label{tab:random_ant}
\end{table*}

\begin{table*}[htb]
    \centering
    \caption{Parameter Randomization details for Halfcheetah}
    \resizebox{\linewidth}{!}{
    \begin{tabular}{c|c|c|c|c}
        System Parameter & Default Value & Training Set & Validation Set & Test Set \\
        \hline\hline
        \rule{0pt}{8pt}
        torso mass & 6.360~kg & & & \\
        \rule{0pt}{8pt}
        two foot mass & 1.069~kg & & & \\
        \rule{0pt}{8pt}
        two thigh mass & 1.535 & & & \\
        \rule{0pt}{8pt}
        two shin mass & 1.581 & $\text{default} \times [0.8, 1.2]$ & $\text{default} \times [0.5, 1.5]$ & $\text{default} \times [0.2, 2.0]$ \\
        \rule{0pt}{8pt}
        two foot friction & 0.4 & & & \\
        \rule{0pt}{8pt}
        front shin joint lower limit & -1.2 & & & \\
        \rule{0pt}{8pt}
        back shin joint lower limit & -0.785 & & &
    \end{tabular}
    }
    \label{tab:random_half}
\end{table*}

\section{Implementation Details and Hyperparameters}
\label{apd:hyper}

For the implementation of ADP, we separate the task sampling module from the policy training. After system parameters selection, we create 10 workers to rollout with each environment in parallel via Ray~\cite{moritz2018ray}, and transitions are gathered for policy training.

For PPO algorithm, we use the same hyperparameters for all experiments. The learning rate is $3\times10^{-4}$, the $\gamma$ is $0.99$, and the $\lambda$ of GAE is $0.95$. The policy ratio clip is $0.25$, the temperature coefficient is $0.1$, the epoch number is $20$, and the batch size is $256$. Each worker produces 5 trajectories for each iteration. 

For the network architecture, the policy network and value function network both have two hidden layers of 64 and 128 neurons respectively. We adopt orthogonal parameter initialization for value function networks and uniform parameter initialization for policy networks. The multi-variant Gaussian distribution with fixed standard deviation is adopted as the policy output. We use $0.4$ as the fixed standard deviation for all environments.

For the system parameter sampling in ADP, we sample 30 parameters from a predefined randomization space and 10 parameters from the replay buffer. The capacity of the replay buffer is 40. After parameter evaluation, we select the best 10 parameters for later rollouts. We use $0.01$ as the bin width (subject to the scales) for the randomization space discretization. 

\section{Baseline Details}
\label{apd:baseline}
For a fair comparison, we implement all baselines with PPO backbone, and the network architectures are the same across all baselines. The explicit hyperparameters adopted in PPO are clarified in \ref{apd:hyper}. For each algorithm, we evaluate the policy in the validation set periodically, and the policy model which performs best in the validation set is chosen for the latter comparison in the test set. Detailed implementation and extra hyperparameters are described below:

\textbf{UDR}: The only difference between ADP and UDR is the task sampling approach. The system parameters are uniformly sampled from the predefined randomization space in UDR. We sample 10 system parameters at each iteration for the rollouts.

\textbf{RARL}: RARL trains robust policy through adversarial training against an adversary policy which adds external force to specific bodies of the robotics. The main policy~(also called protagonist) and adversary policy obtain inverse reward values at each step, and two policies are trained alternatively to stabilize training. For the adversary agent which applies external forces to specific bodies of the robotics, we follow the official implementation: the observation space of the adversary is the same as the protagonist in each environment, the action space of the adversary varies across environments~(2D forces on the foot of Hopper, 4D forces on both feet of Walker, 6D forces on the torso and both feet of Halfcheetah, 8D forces on all feet of Ant). The maximum force is set to 1.0 for each environment. The training epochs of the protagonist and the adversary are both set to 1.

\textbf{EPOpt}: At each iteration, EPOpt samples several trajectories from different systems and define a cumulative reward threshold at $\epsilon$-percentile of cumulative rewards of all the trajectories. The policy is trained using trajectories whose cumulative rewards are lower than the threshold. To maintain sample efficiency, we use $\epsilon = 1$ at first $\frac{N}{2}$ iterations, and use $\epsilon = 0.1$ at the final $\frac{N}{2}$ iterations ($N$ is the number of total iterations) following the original paper. Since we only compare the zero-shot performance in the test set, we discard the adaptation of the source domain distribution in the original EPOpt.









\end{document}